\documentclass[lettersize, journal, twoside]{IEEEtran}
\usepackage{amsmath, amsfonts}
\usepackage{amsthm}

\usepackage{bm}
\usepackage{array}
\usepackage{graphicx}
\usepackage[caption=false, font=normalsize, labelfont=sf, textfont=sf]{subfig}
\usepackage{textcomp}
\usepackage{stfloats}
\usepackage{url}
\usepackage{verbatim}
\usepackage{xcolor}
\usepackage{graphicx}
\graphicspath{{./}{./TRO2025/Figure/}{../TRO2025/Figure/}}
\usepackage{cite}

\usepackage{booktabs} 
\usepackage{amssymb}

\usepackage{xcolor}       
\usepackage{pifont}       

\usepackage[linesnumbered, ruled, vlined]{algorithm2e}
\usepackage{threeparttable}

\newcommand{\cmark}{\textcolor{green}{\ding{51}}}  
\newcommand{\xmark}{\textcolor{red}{\ding{55}}}    

\DeclareUnicodeCharacter{2009}{ } 
\begin{document}

\markboth{%
}{%
  Nie and Li \MakeLowercase{\textit{et al.}}:  Unified Linear Parametric Map Modeling and Perception-aware Trajectory Planning for Mobile Robotics%
}

\author{Hongyu~Nie$^{1}$,  Xu~Liu$^{3}$,  Zhaotong Tan$^{2}$,  Sen Mei$^{4}$,  Wenbo~Su$^{1}$%
\thanks{Corresponding authors:  Hongyu Nie.}%
\thanks{$^{1}$Hongyu Nie and Wenbo Su are with the School of Information Science and Engineering,  Shenyang University of Technology,  Shenyang 110870,  China; {\tt\footnotesize niehongyu@sia.cn; wudmfx8@163.com}.}%
\thanks{$^{2}$Zhaotong Tan is with the College of Information Science and Engineering and the State Key Laboratory of Synthetic Automation for Process Industries,  Northeastern University,  Shenyang 110819,  China and intern with Woozoom Technology Company Ltd.,  Shenyang 110179,  China; {\tt\footnotesize \{2400904\}@stu.neu.edu.cn}.}%
\thanks{$^{3}$Xu Liu is with the State Key Laboratory of Robotics,  Shenyang Institute of Automation,  Chinese Academy of Sciences,  Shenyang 110016,  China,  and also with the University of Chinese Academy of Sciences,  Beijing 100049,  China; {\tt\footnotesize liuxu172@mails.ucas.ac.cn}.}%
\thanks{$^{4}$Sen Mei is with Woozoom Technology Company Ltd.,  Shenyang 110179,  China; {\tt\footnotesize mason368@126.com}.}%
}

\title{Unified Linear Parametric Map Modeling and Perception-aware Trajectory Planning for Mobile Robotics}

\maketitle

\begin{abstract}
\textcolor{black}{Autonomous navigation in mobile robots,  reliant on perception and planning,  faces major hurdles in large-scale,  complex environments. These include heavy computational burdens for mapping,  sensor occlusion failures for UAVs,  and traversal challenges on irregular terrain for UGVs,  all compounded by a lack of perception-aware strategies. To address these challenges,  we introduce Random Mapping and Random Projection (RMRP). This method constructs a lightweight linear parametric map by first mapping data to a high-dimensional space,  followed by a sparse random projection for dimensionality reduction. Our novel Residual Energy Preservation Theorem provides theoretical guarantees for this process,  ensuring critical geometric properties are preserved. Based on this map,  we propose the RPATR (Robust Perception-Aware Trajectory Planner) framework. For UAVs,  our method unifies grid and Euclidean Signed Distance Field (ESDF) maps. The front-end uses an analytical occupancy gradient to refine initial paths for safety and smoothness,  while the back-end uses a closed-form ESDF for trajectory optimization. Leveraging the trained RMRP model's generalization,  the planner predicts unobserved areas for proactive navigation. For UGVs,  the model characterizes terrain and provides closed-form gradients,  enabling online planning to circumvent large holes. Validated in diverse scenarios,  our framework demonstrates superior mapping performance in time,  memory,  and accuracy,  and enables computationally efficient,  safe navigation for high-speed UAVs and UGVs.}
\end{abstract}

\section{\textcolor{black}{Introduction}}

\begin{figure}[t]
	\centering 
	\includegraphics[width=0.5\textwidth]{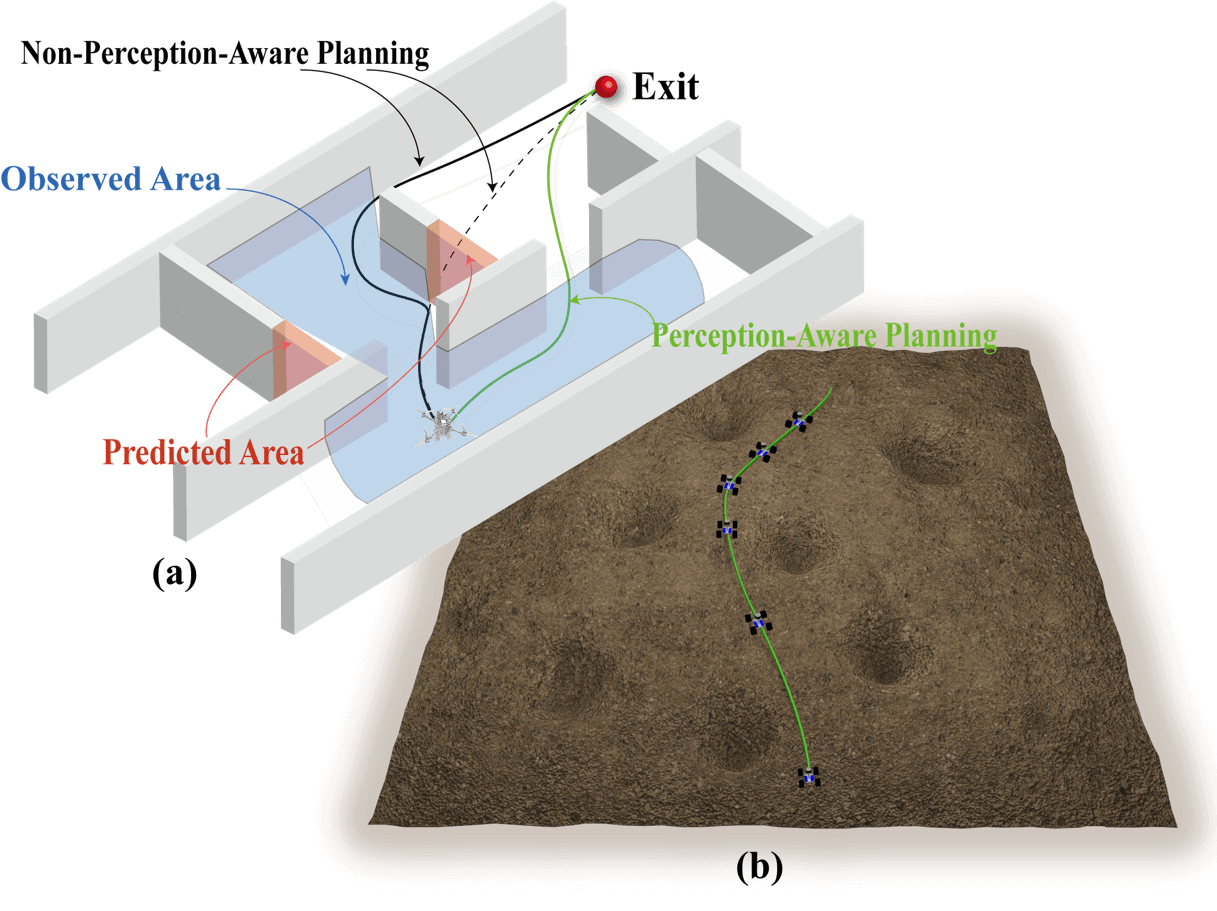}
	\caption{\textcolor{black}{Applications of the RMRP perception-aware planning framework. (a) For UAVs,  obstacle completion in sensor blind-spots enables proactive avoidance and safer,  smoother trajectories. (b) For UGVs,  analytical terrain gradients serve as a risk cost for trajectory optimization to safely circumvent depressions. We will soon release a video demonstration to provide more information about the experiments.}
}
	\label{introduction}
\end{figure}

\IEEEPARstart{A}utonomous navigation for mobile robots has demonstrated efficient performance in both indoor and outdoor settings. To accomplish complex tasks,  such as search and rescue \cite{queralta2020collaborative},  autonomous driving \cite{han2023efficient},  and industrial inspection \cite{xing2023autonomous}. Mobile robots require robust and reliable navigation systems,  which fundamentally consist of two components:  environmental perception and motion planning. However,  enabling robust autonomous navigation in large-scale,  cluttered environments presents substantial challenges,  mainly in constructing suitable maps and developing adaptive motion planning algorithms.

 Navigation Maps:   Occupancy grid map can provide unknown,  known,  and occupied environmental areas,  which is crucial for motion planning. ESDF and terrain maps can provide gradient information for trajectory optimization. However,  these map representations are highly resolution-sensitive. In occupancy grid maps,  it face challenges in balancing memory consumption and mapping accuracy. In ESDF and terrain maps,  The gradient information related to Euclidean distance or elevation values is obtained through interpolation calculation by resolution.  Critically,  these maps struggle to balance memory consumption with mapping precision,  especially for large-scale and unstructured environments.

 Motion Planning:   Despite significant advancements in robotic motion planning,  substantial challenges persist in complex scenarios. As illustrated in \textcolor{black}{\textcolor{black}{Fig.~\ref{introduction}}},  these include 1) High-speed navigation of quadrotors in unknown environments. 2) The car-like robots safely and quickly avoids these large pits in the uneven mountainous.  The existing planning methods mainly have the following issues. (a) During high-speed flight in unknown environments,  UAVS must generate collision-free trajectories within short time constraints to avoid unforeseen obstacles. Unfortunately,  current methods are difficult to find safe,  smooth,  and continuous trajectories in limited time. (b) Gradient-based obstacle avoidance planning methods rely on appropriate resolution and are difficult to safely and efficiently adapt to different scenarios. (c) Existing methods are unaware of environment perception,  they neglecting key information on the map and failing to integrate it well with planning.

 To address these challenges,  we introduce a Lightweight Linear Parameter Map constructed via RMRP.  Based on this continuous,  parametric map representation,  we propose RPATR (Robust Perception-Aware Trajectory Planner),  a novel perception-aware planning framework for mobile robots in large-scale,  cluttered,  and unknown environments.


\textcolor{black}{Robotic environment modeling (e.g.,  occupancy mapping,  terrain representation) is crucial for autonomous navigation. Although methods like Random Mapping Method (RMM) \cite{liu2022unified} can efficiently project data into high-dimensional spaces,  the resulting representations impose significant computational and storage burdens,  making dimensionality reduction a primary challenge. The JL Lemma,  while providing a basis for dimensionality reduction via approximate pairwise distance preservation \cite{Johnson1984,  clarkson2008manifold},  fails to ensure norm preservation for arbitrary vectors or subspaces. This limits its utility in regression or classification tasks requiring residual geometry preservation (e.g.,  $\mathbf{A}x - \mathbf{y}$) \cite{garivier2024sparsity,  klopotek2017mljl},  motivating stronger subspace-preserving techniques. OSEs address this by ensuring approximate norm preservation for k-dimensional subspaces,  accelerating linear algebra \cite{cohen2016ose,  nelson2013lower,  2012osnap}. However,  existing OSEs are often task-agnostic,  limiting effectiveness for domain-specific structures (e.g.,  classification boundaries,  regression residuals) \cite{slawski2017pcr,  li2017distance,  li2017rip}.}

\textcolor{black}{To overcome these limitations,  we propose a method combining RMRP's efficiency with task-adaptive subspace embedding,  designing projections to explicitly preserve geometric features within task-relevant subspaces,  ensuring both norm and subspace relationship fidelity. Our method outperforms RMM in classification or regression. Detailed contributions follow.}

1) \textcolor{black}{We propose a \textbf{Residual Energy Preservation Theorem} that provides a quantitative $O(\epsilon)$-level bound on the projection residual. By synergizing subspace embedding,  angle-leakage control,  and sparse concentration inequalities,  this theorem fills a critical gap in traditional random projection theory.}

2) \textcolor{black}{This theorem provides end-to-end guarantees for applying SRP to classification and regression tasks,  preserving model stability and generalization. It yields a precise dimension-selection formula,  $k=\Theta(p\log(p/\delta)/\varepsilon^2)$,  and reduces computational cost from $O(dk)$ to $O(s\, k)$.}



3) \textcolor{black}{Based on the above,  we introduce a \textbf{unified parametric mapping framework} using RMRP. We reformulate occupancy grid mapping as a classification task to yield a continuous,  lightweight,  and memory-efficient map,  which is central to all subsequent planning applications.}



4) \textcolor{black}{We exploit the map's \textbf{continuous and differentiable properties} to enhance front-end path searching. By incorporating the analytical occupancy gradient into a novel cost function,  we refine initial paths from planners like Kinodynamic $A^*$ for significantly improved safety and smoothness.}


5) \textcolor{black}{To complete the planning pipeline,  we apply the RMRP framework to \textbf{back-end trajectory optimization}. By modeling the ESDF as a regression task,  we obtain a closed-form,  interpolation-free representation that provides highly accurate and continuous gradients for the optimizer.}



6) \textcolor{black}{Our framework enables \textbf{predictive,  perception-aware planning} by using a learned RMRP model to complete scenes,  inferring occupancy in unobserved areas based on learned geometric priors,  allowing the planner to proactively generate safer and more efficient trajectories for high-speed flight.}


7) \textcolor{black}{Finally,  by \textbf{extending} the RMRP framework to \textbf{UGV navigation},  we demonstrate its versatility. We model uneven terrain as a regression task and leverage its analytical gradient for robust,  terrain-aware trajectory optimization that safely avoids hazards such as large depressions.}

\section{Related Work}
\subsection{Navigation Maps}
A desirable map for autonomous navigation can accommodate high-quality environment model and good performance in motion planning. Occupancy grid maps and ESDF maps are commonly used for path searching and trajectory optimization,  respectively.
\textbf{Occupancy maps} are a common and classic map for motion planning in \cite{hornung2013octomap},  \cite{chen2016online1}. The construction process of occupancy grid maps is very simple and clear. First,  assumes priori discretization of the space into grid cells,   and then estimates the occupancy state( free or occupied). For the grid cell,  there are many different storage forms,  such as the octree-based \cite{hornung2013octomap},  hash table-based \cite{niessner2013real} and uniform grid-based methods \cite{zhou2019robust}. However,  as the storage consumption scales linearly with map size,  these methods have poor performance in large-scale environments. In order to obtain the occupancy state (free or occupied),   the classic estimate method is the fusion-based \cite{doherty2019learning} \cite{srivastava2018efficient} \cite{o2012gaussian}. Although these methods are simple and efficient,  fusion-based methods need to store the predicted results for future state estimation,  leading to challenges in computing and storage under limited resources. In addition,  some novel method-based deep learning \cite{bauer2019deep} \cite{senanayake2017deep} can infer the occupancy state from the original points in an end-to-end manner. Typically,  this kind of method needs more training time.

\textbf{ESDF maps} Based on the global occupancy map,  the \textbf{ESDF map} (local map) is built based on Euclidean distance \cite{felzenszwalb2012distance} \cite{lau2010improved} and  gradient information by interpolation \cite{oleynikova2017voxblox} \cite{han2019fiesta}. The ‘Voxblox’ method and FIESTA are proposed in \cite{oleynikova2017voxblox} \cite{han2019fiesta} to build the ESDF map incrementally. \cite{chen2022gpu} use GPU programming techniques to fast construct ESDF map for motion planning. Recently,  \cite{pantic2022sampling} used Neural Radiance Fields to construct ESDF map. These gradient-based planning methods can use the ESDF map for obstacle avoidance \cite{zhou2019robust} \cite{oleynikova2016continuous} \cite{tordesillas2019faster}. However,  these methods rely on resolution,  so they need to constantly adjust the appropriate resolution to adapt to different scenes,  which,  in return,  limits the performance of autonomous navigation in complex field environments.

\textbf{Terrain maps}
Currently,  there are many classic and commonly used methods for terrain modeling. Traditional methods,  based on different storage formats,  such as point cloud maps \cite{nuchter20076d} \cite{lalonde2006natural},  elevation maps \cite{lacroix2002autonomous}\cite{triebel2006multi},  and grid-based maps \cite{hornung2013octomap},  are used for terrain representation. For large-scale terrain modeling,  when fine resolution is required,  these methods occupy huge storage space. In addition,  these methods do not consider the spatial relevance of terrain so that they are not suited for motion planning on uneven terrain. To address these limitations,  Gaussian processes (GPs) \cite{ounpraseuth2008gaussian} offer an elegant solution as a powerful Bayesian non-parametric regression technique. Leveraging probabilistic inference capabilities \cite{liu2021multiresolution},  GPs generate terrain models at arbitrary resolutions while encoding spatial structures through covariance functions. However,  existing GP implementations exhibit limited generalization capabilities,  which hinder their practical deployment. Critically,  their O(N³) computational complexity with training data size renders them infeasible for large-scale terrain modeling.

\subsection{Path search}
Path finding can be divided into two classical methods:  graph searching-based and sampling-based planning.

\textbf{Searching‐based methods}:  Search-based algorithms involve discretizing the configuration space and converting the problem to graph searching. Dijkstra's algorithm \cite{dijkstra2022note}  is widely used for finding the shortest path on the entire graph. A*  \cite{hart1968formal}is the most commonly used heuristic search algorithm and it is an extension of Dijkstra's algorithm. A* allows for fast node search by using heuristics. The efficiency of the A* search depends on the heuristics designed. A good heuristic will lead to faster convergence and fewer node expansions. It is suitable for \textbf{search space} on the global prior graph,  such as an occupancy grid map. Many improved methods are based on A*,  such as Anytime A* \cite{hansen2007anytime},  Jump Point Search (JPS) \cite{harabor2011online},  and hybrid A* \cite{dolgov2010path}. Anytime A* algorithms attempt to provide a first sub-optimal path quickly and continually improve the optimality of the path. JPS identifies "jump points" that allow it to skip over large free-space nodes of uniform-cost grid maps. By pruning nodes that don't contribute to finding the optimal path,  JPS can significantly reduce the \textbf{search space} compared to traditional A*. Hybrid A* considers the point-mass model of a multi-rotor in searching a dynamically feasible path.  Methods using the state-lattice \cite{liu2017search} construct a search space consisting of motion primitives. The search space is a graph that represents the possible states and transitions of the multi-rotor in a discretized space. 

\subsection{ Perception-Aware Planning }
In recent decades,  various approaches on perception-aware planning have been proposed for mobile robotics in unknown and cluttered environments. For the fast flight of UAVs on unknown environments, 
Chen et al. \cite{ chen2024apace} introduced a novel decomposable visibility model within a two-stage framework for efficient perception-aware trajectory generation. However,  this method overlooks the observation of unknown areas.
\cite{zhou2021raptor} Using active heading adjustment to perceive unknown areas due to obstacles occlusion or limited sensor field of view,  increasing planning time,  and making it difficult to maintain high speed. Now,  some learning-based approaches \cite{wang2021learning} \cite{song2023learning} are used for perception-aware planning,   by utilizing predicted maps to enable global planning. Unfortunately,  these methods demand substantial training data and exhibit poor performance in real-time quality. For the car-like robots on uneven terrain,  
\cite{krusi2017driving} proposed a trajectory planning method that considers the traversability cost of terrain. However,  this method requires a large amount of sampling and iterative computation,  which is not suitable for online tasks. \cite{jian2022putn} proposed a plane-fitting based uneven terrain navigation framework. The initial path is obtained by informed-RRT* in the front end, and integrate the terrain cost function into NMPC to generate a local trajectory. Unfortunately,  complexity of the NMPC problem will rise dramatically as the planning horizon becomes longer.


\subsection{Dimensionality reduction}

\textcolor{black}{Dimensionality reduction is essential in robotics and machine learning for scaling algorithms while preserving geometric properties. This section summarizes key developments in this area,  focusing on the Johnson–Lindenstrauss (JL) lemma,  oblivious subspace embeddings (OSE),  and efficient projection methods. 
The JL lemma ensures that any $N$ points in high-dimensional space can be embedded into $m = O(\varepsilon^{-2} \log N)$ dimensions while preserving pairwise distances. Sparse and binary variants reduce computation and storage costs. Achlioptas proposed binary projections with $\{+1,  0,  -1\}$ entries,  while Matoušek developed sparse constructions that maintain accuracy with fewer non-zeros per column \cite{matousek2008jl,  garivier2024sparsity,  fedoruk2018dimensionality}. For preserving norms of all vectors in a $k$-dimensional subspace,  OSEs provide stronger guarantees. Sarlós introduced this concept for regression and low-rank approximation. Later work established tight lower bounds for sparse embeddings and showed they can be constructed with optimal projection dimension $m = O((k + \log(1/\delta)) / \varepsilon^2)$,  independent of the dataset size \cite{cohen2016_ose,  nelson2013lower}. Ailon and Chazelle proposed the Fast Johnson–Lindenstrauss Transform (FJLT),  reducing projection time to $O(d \log d + m \log m)$ via structured transforms and randomization \cite{ailon2006fjl}. Liberty et al. later extended this with lean Walsh transforms for improved efficiency in dense and streaming settings \cite{liberty2008fast}. While current methods provide fast and memory-efficient embeddings,  they lack theoretical guarantees for preserving residual energy orthogonal to the principal data subspace. This property is critical for stabilizing downstream tasks and motivates our more comprehensive analysis.}


\section{\textcolor{black}{system overview}}
\begin{figure*}[t]
	\centering 
	\includegraphics[width=1.0\textwidth]{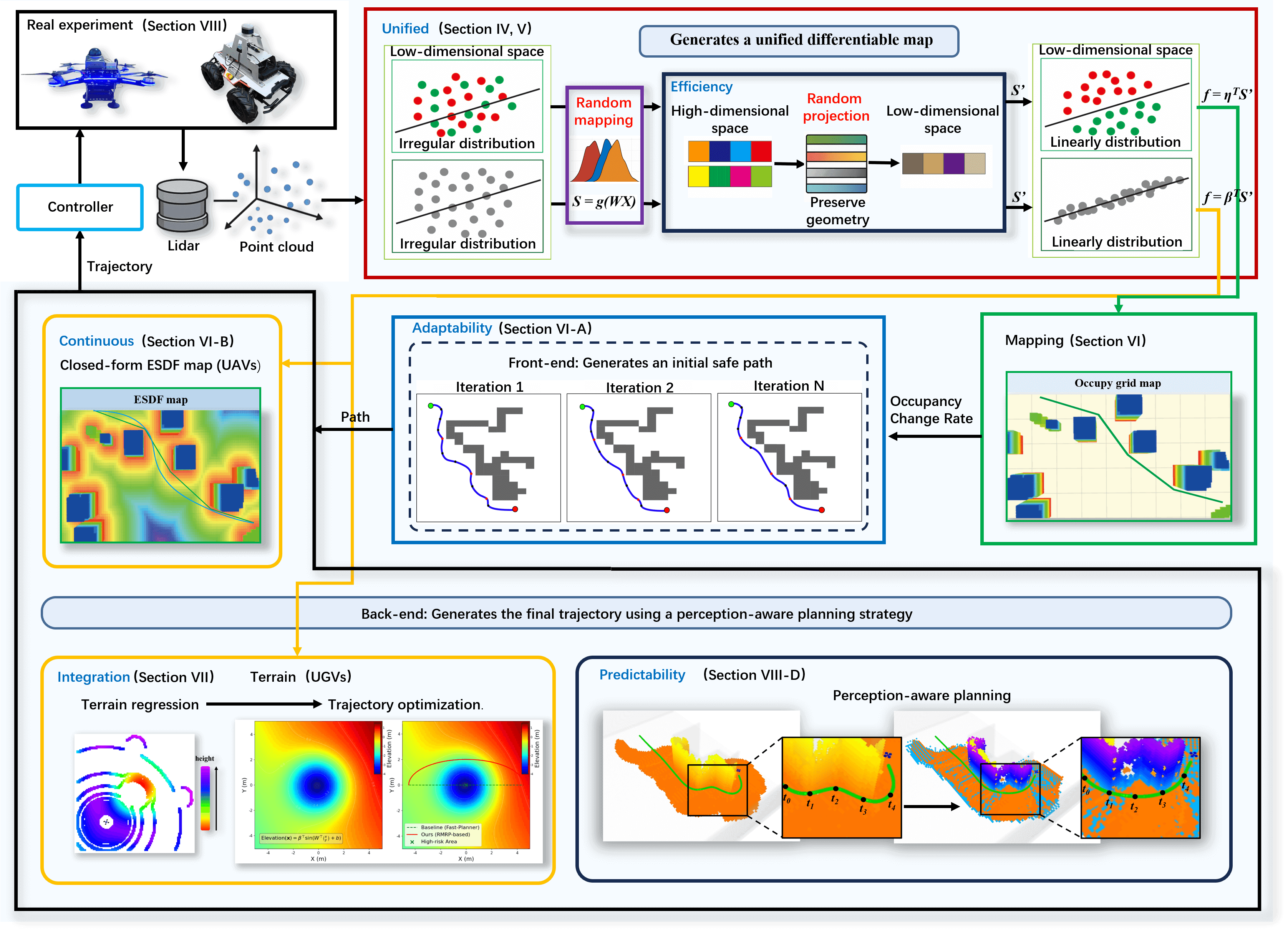}
	\caption{\textcolor{black}{The proposed LPATR framework for autonomous flight integrates environmental perception and motion planning through a unified,  lightweight,  and continuous parametric model. The process begins with raw point cloud inputs,  which are transformed into a unified differentiable map via a RMRP module. This module employs high-dimensional mapping and efficient random projection to convert the data into a continuous,  differentiable parametric environment model. Serving as the central link between perception and planning,  this model supports both the front-end path searching and the back-end trajectory optimization. The framework's modular design efficiently integrates distinct planning strategies for Unmanned Aerial Vehicles (UAVs) and Unmanned Ground Vehicles (UGVs),  enabling end-to-end perception-aware navigation.}
}
	\label{Framework}
\end{figure*}

\textcolor{black}{
The proposed navigation system is illustrated in \textcolor{black}{Fig.~\ref{Framework}}. It takes raw point clouds as input and converts them into a unified,  lightweight,  continuous parametric model of the environment. First,  the Random Mapping method projects the input data into a high-dimensional feature space where it becomes linearly separable (Section IV). Second,  a sparse random projection reduces the dimensionality to ensure computational efficiency while preserving key geometric structures (Section V). This process generates a unified differentiable map (Section VI),  which is then used in a two-stage planning pipeline. The front-end finds an initial path,  which is then refined using the map's analytical gradient to enhance safety and smoothness (Section VI-A). The back-end then employs a perception-aware planning strategy to generate the final trajectory. It leverages the map's outputs—such as the closed-form ESDF for UAVs and the analytical terrain gradient for UGVs—for efficient,  risk-aware optimization (Section VI-B,  VII). Crucially,  this strategy uses the model's predictive \textcolor{black}{ability} for real-time completion of occluded obstacles,  enabling the robot to proactively navigate around unseen hazards (Section VIII-D).}

\section{Random Mapping}

\textcolor{black}{
This section reviews the theoretical foundations of the Random Mapping Method (RMM). While the kernel trick is effective for nonlinear problems,  constructing a suitable kernel function is often difficult. To address this issue,  a recent work proposed the RMM method~\cite{liu2022unified},  which projects data from a low-dimensional input space into a high-dimensional feature space where it becomes approximately linearly separable. This property enables the use of simple linear models to solve complex problems.}

\subsection{Random Mapping Concept}

\textcolor{black}{
The RMM~\cite{liu2022unified} generates a high-dimensional Random Mapping set (RM set),  $S$,  from an input set X using the formula: 
\begin{equation}
S = g\left(WX\right)
\end{equation}
The components are defined as follows: 
$X \in \mathbb{R}^{N \times L}$ is the input set containing L samples.
$W \in \mathbb{R}^{M \times N}$ is a random weight matrix whose elements are drawn from a probability distribution (e.g.,  a uniform distribution). The parameter $M$ is the \textbf{random mapping dimension}.
$g(\cdot)$ is a nonlinear activation function (e.g.,  sine) applied element-wise.
$S \in \mathbb{R}^{M \times L}$ is the resulting RM set,  where each column is a new high-dimensional feature vector.}

\textcolor{black}{
The objective of this process is to generate an RM set S with a \textbf{linear property},  which is essential for enabling linear models to fit the data~\cite{liu2022unified}. This makes the RMRP framework readily applicable to both classification and regression tasks~\cite{liu2022unified}.}

\subsection{Feasibility and Existence Analysis}

\textcolor{black}{
The theoretical validity of RMRP is established by proving the following core claim~\cite{liu2022unified}: }

\textcolor{black}{
\textbf{Claim: } For any target vector T and any arbitrarily small positive real number $\epsilon > 0$,  there exists a natural number P,  such that when the random mapping dimension M satisfies $M > P$,  at least one linear model parameter $\beta$ exists that satisfies the condition $||\beta^T S - T|| < \epsilon$.}

\textcolor{black}{
The proof of this claim rests upon two key lemmas also presented in~\cite{liu2022unified}: 
1.  The first lemma establishes that when M is sufficiently large,  the RM set S almost surely achieves full column rank. This implies that a model capable of a perfect fit (zero error) exists with a probability of 1.
2.  The second lemma demonstrates that the optimal approximation error is monotonically non-increasing as the mapping dimension M grows.}

\textcolor{black}{
Together,  these lemmas and the monotone convergence theorem provide a solid theoretical guarantee for the RMRP method. This ensures that a linear model can approximate the target function to any desired degree of accuracy by using a sufficiently large mapping dimension M.}

\subsection{Iterative Solution of $\beta$} 
\textcolor{black}{To ensure the model can be learnt online,  we employ the AdamW optimizer \cite{loshchilov2019adamw}. In contrast to standard SGD,  this method utilizes adaptive learning rates and a decoupled weight decay mechanism to improve convergence and generalization performance \cite{kingma2015adam,  loshchilov2019adamw}. The iterative update of the weight vector $\beta$ at each time step $t$ is given by}

\begin{equation}
\beta_t = \beta_{t-1} - \gamma \left( \frac{\hat{m}_t}{\sqrt{\hat{v}_t} + \epsilon} + \lambda \beta_{t-1} \right)
\end{equation}

\textcolor{black}{where $\eta$ is the learning rate and $\lambda$ is the weight decay factor. The terms $\hat{m}_t$ and $\hat{v}_t$ represent the bias-corrected first and second moment estimates,  which are computed from the gradient using exponential decay rates $\beta_1$ and $\beta_2$ \cite{kingma2015adam}. The term $\epsilon$ is a small constant added for numerical stability. The time complexity of this optimizer is linear with respect to the model parameters $M$. While the per-iteration cost is comparable to that of SGD,  AdamW's primary efficiency gain stems from a faster convergence rate,  often reducing the total number of iterations required \cite{kingma2015adam,  loshchilov2019adamw}. This allows the model to be updated efficiently online,  fitting the overall unified framework.}

\section{Random Projection}

\textcolor{black}{
In specific application scenarios,  the combination of RM and Sparse Random Projection (SRP) may not be fully supported by the JL lemma alone. While the JL lemma ensures approximate preservation of pairwise distances during dimensionality reduction,  it does not guarantee the retention of structural properties critical to downstream tasks. The RM approach maps input data into a high-dimensional space $H_{RM} = \{ \bm{S} | \bm{S} = g(W\bm{x}),  \bm{x} \in X \}$,  where nonlinear activation functions induce an approximately linear distribution that facilitates linear regression and classification tasks. Subsequently,  SRP reduces the dimensionality from $H_{RM}$ to a lower-dimensional space $H_{SRP} = \{ \bm{S'}|\bm{S'} = R \bm{S},  \bm{S} = g(W\bm{x}),  \bm{x} \in X \}$,  aiming to reduce computational cost. However,  the JL lemma focuses solely on distance preservation and overlooks essential structural characteristics introduced by RM,  such as the geometry and orientation of linear subspaces and the residual distributions from these subspaces. These attributes are crucial for the performance of linear models and therefore must be preserved to maintain effectiveness in regression and classification tasks.
}

\begin{figure}[h]
	\centering 
	\includegraphics[width=0.49\textwidth]{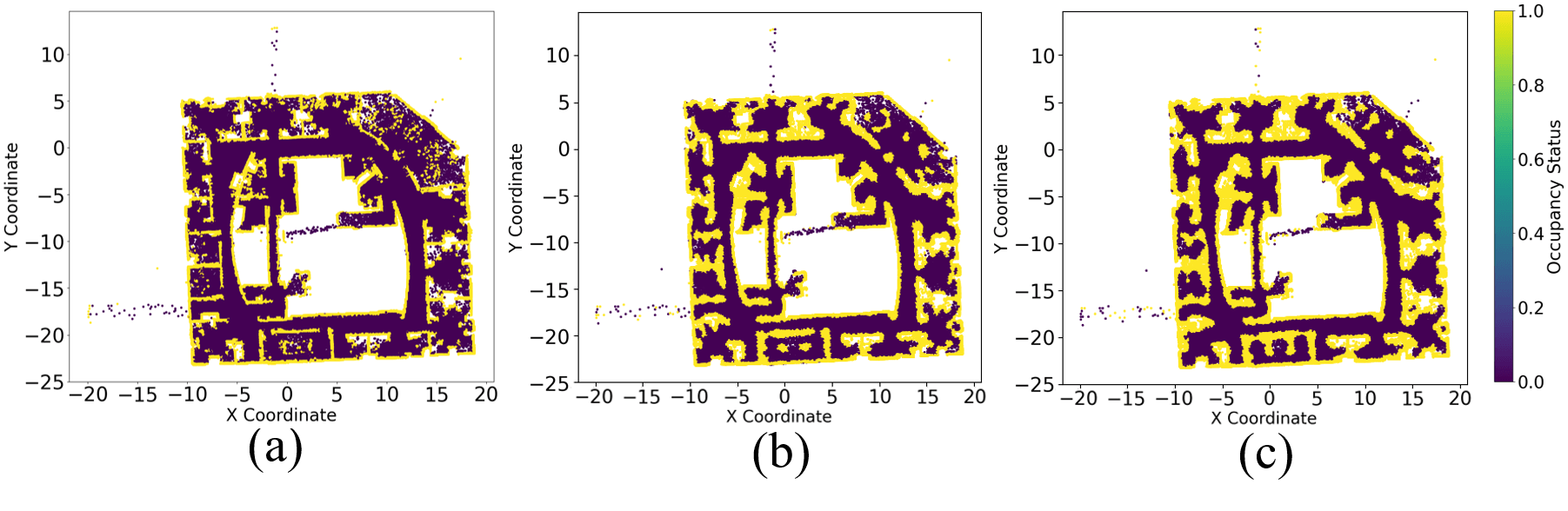   }
	\caption{\textcolor{black}{presents occupancy grid maps generated on Dataset 1 using three methods:  (a) Ground Truth,  (b) RMM,  and (c) RMRP. Qualitative analysis indicates that (c) exhibits strong visual consistency with (a) and (b),  validating its capacity to preserve high classification accuracy while markedly reducing model dimensionality.}
 } 
	\label{classify}%
\end{figure}

\begin{figure}[h]
	\centering 
	\includegraphics[width=0.49\textwidth]{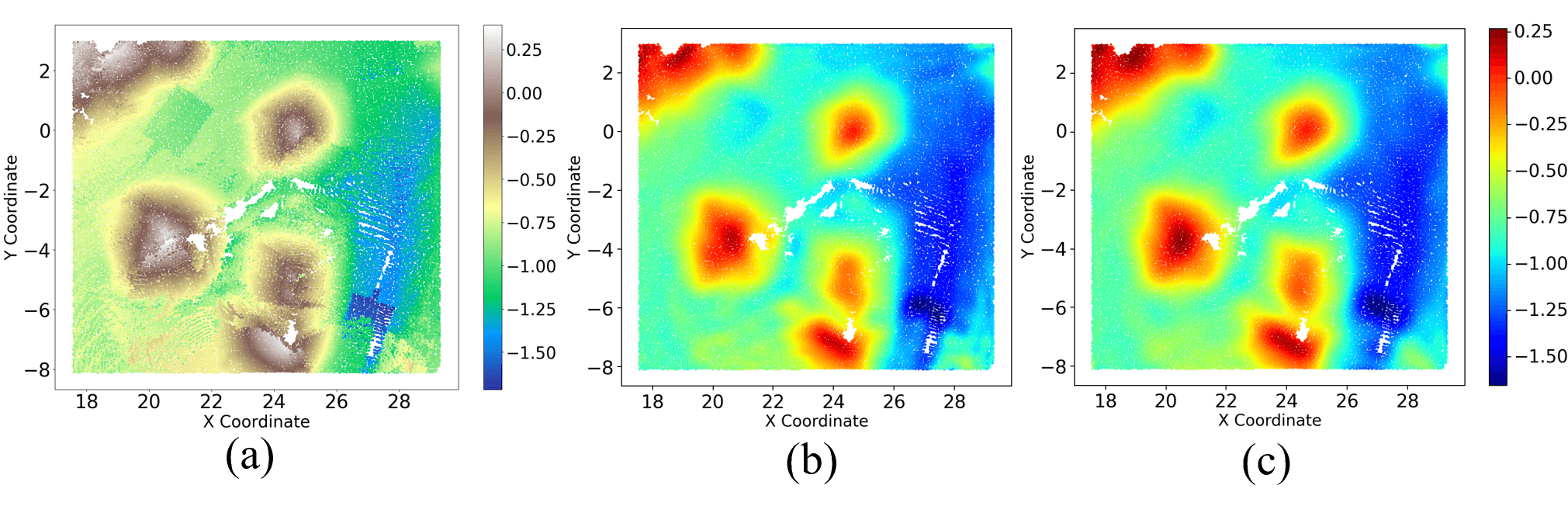   }
	\caption{\textcolor{black}{validates the fidelity and efficiency of RMRP in terrain regression. The figure compares the modeling results of (a) Ground Truth,  (b) RMM,  and (c) RMRP. RMRP accurately reconstructs key terrain features,  such as geometry and gradients,  with lower overhead,  providing reliable gradient information for subsequent planning.}
 } 
	\label{regression}%
\end{figure}

\begin{figure}[h]
	\centering 
	\includegraphics[width=0.49\textwidth]{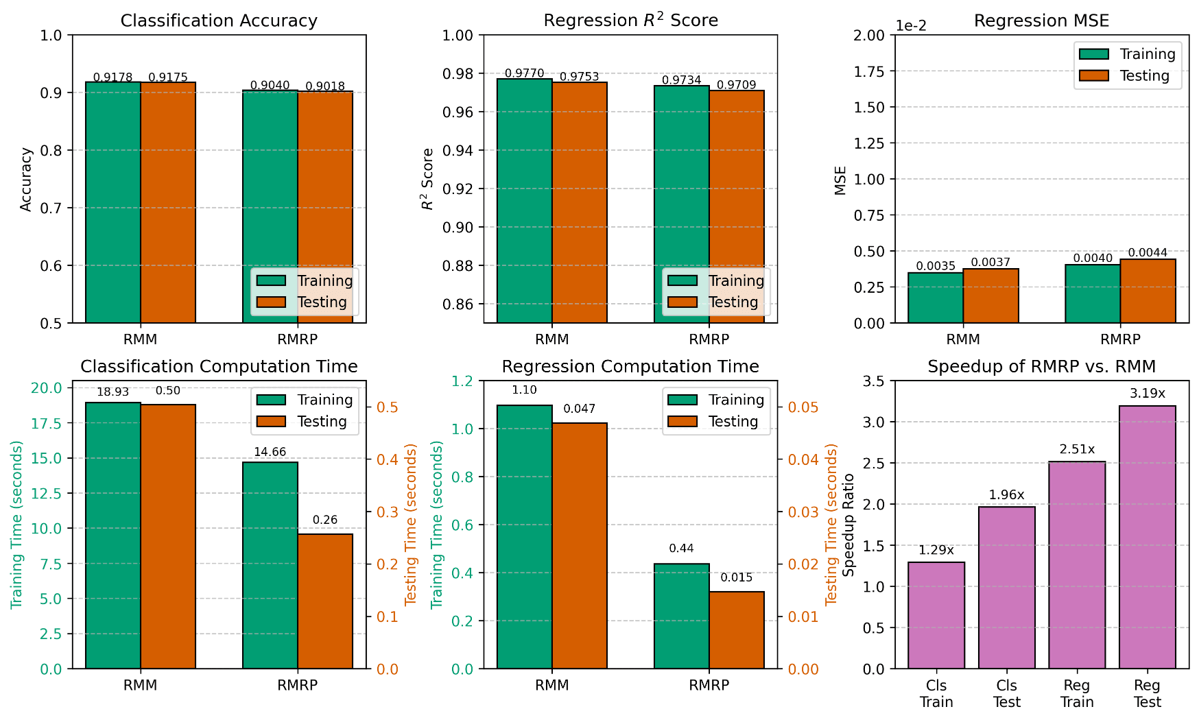   }
	\caption{\textcolor{black}{quantitatively evaluates the performance of RMM and RMRP in classification and regression tasks. The results confirm that RMRP achieves comparable classification accuracy and regression metrics (R²,  MSE) to RMM,  while drastically reducing training and testing computation time. This significant speedup validates the effectiveness of sparse random projection as a powerful acceleration technique for large-scale mapping and planning without sacrificing prediction performance.}
 } 
	\label{data}%
\end{figure}

\textcolor{black}{
This section describes the subspace structure of high-dimensional feature vectors after they are lifted by RM,  and presents the construction of SRP for dimensionality reduction,  along with the necessary and sufficient conditions that guarantee subspace isometric embedding.}

\begin{enumerate}
  \item \textbf{\textcolor{black}{Subspace Decomposition.}}  
  \textcolor{black}{
    Let $\mathbf{x}\in\mathbb{R}^d$ denote a high-dimensional feature vector obtained by lifting an original input via RM. Suppose that $\mathbf{x}$ admits a unique orthogonal decomposition with respect to a linear subspace $\mathcal{S}\subset\mathbb{R}^d$ of dimension $p$: }
    \begin{equation}
      \mathbf{x} \;=\; \mathbf{x}_{\mathcal{S}} + \mathbf{x}_{\perp},  
      \quad \mathbf{x}_{\mathcal{S}} = P_{\mathcal{S}}\, \mathbf{x},  
      \;\; \mathbf{x}_{\perp} = P_{\mathcal{S}}^{\perp}\, \mathbf{x}, 
    \end{equation}
    \textcolor{black}{
    where $P_{\mathcal{S}}$ is the orthogonal projector onto $\mathcal{S}$ and $P_{\mathcal{S}}^{\perp}$ projects onto its orthogonal complement $\mathcal{S}^{\perp}$ \cite{Johnson1984}.}

    \item \textbf{\textcolor{black}{Motivation:  Preserving Linear Separability.}}  
      \textcolor{black}{
      By the RMM mapping,  original data become almost linearly separable in a high-dimensional feature space (cf.\ \textcolor{black}{Fig.~\ref{method}~}).  
      To carry this separability through to the reduced space,  we must not only preserve pairwise distances (the JL guarantee)  
      but also maintain the geometry and orientation of the learned subspace and control the energy of the orthogonal-complement “noise.”  
      The following SRP analysis ensures exactly that.}

  \item \textbf{\textcolor{black}{Sparse Random Projection (SRP).}} 
  \textcolor{black}{
    In order to reduce computational complexity without sacrificing essential information,  the high-dimensional vector $\mathbf{x}$ is mapped to a lower-dimensional space via SRP. Let $\mathbf{R}\in\mathbb{R}^{k\times d}$ be a $\{0, \pm1\}$-valued random matrix constructed according to Achlioptas’ database-friendly scheme,  with density parameter $1 - s$. Such a construction ensures that the cost of computing $\mathbf{R}\, \mathbf{x}$ scales in proportion to the number of nonzero entries of $\mathbf{x}$,  yielding a time complexity of $O\bigl(\mathrm{nnz}(\mathbf{x})\bigr)$ for matrix–vector multiplication in high-dimensional sparse settings \cite{maillard2012linear, Andoni2024}. The projected vector is then given by}
    \begin{equation}
      \mathbf{x}' \;=\; \mathbf{R}\, \mathbf{x},  
      \quad \mathbf{x}'\in\mathbb{R}^{k}.
    \end{equation}
    \textcolor{black}{
    To further reduce computational cost,  one may employ the Fast Johnson–Lindenstrauss Transform (FJLT) of Ailon and Chazelle,  which achieves a time complexity of $O\bigl(d\log d\bigr)$ \cite{AilonChazelle2009}.}


  \item \textbf{\textcolor{black}{Subspace Isometric Embedding.}}
  \textcolor{black}{
    By leveraging Clarkson–Woodruff’s input‐sparsity subspace embedding result and its optimal dimension lower bound,  we establish that if}
    \begin{equation}
      k \;\ge\; C\, \frac{p\, \log(p/\delta)}{\varepsilon^{2}}
      \label{eq: ose-dim}
    \end{equation}
    \textcolor{black}{
    then,  with probability at least $1-\tfrac{\delta}{2}$ \cite{Clarkson2013, Chenakkod2023},  the random projection $R\in\mathbb{R}^{k\times d}$ satisfies}
    \begin{equation}
      (1 - \varepsilon)\, \|v\| \;\le\;\|\mathbf{R}\, v\|\;\le\;(1 + \varepsilon)\, \|v\|,  
      \quad \forall\, v\in \mathcal{S}\cup \mathcal{S}^{\perp}
    \end{equation}
    \textcolor{black}{
    In other words,  $R$ achieves an $\varepsilon$-subspace embedding,  guaranteeing that the lengths of all vectors in both the target subspace $\mathcal S$ and its orthogonal complement are preserved within a relative error $\varepsilon$.}
\end{enumerate}

\subsection{Projection Decomposition of the Residual Vector}

\textcolor{black}{
Under the assumptions of the previous subsection,  denote by $P_{\mathcal{S}}$ and $P_{\mathcal{S}}^{\perp}$ the orthogonal projectors onto $\mathcal{S}$ and its orthogonal complement in $\mathbb{R}^d$,  respectively. After applying SRP,  one obtains the lower-dimensional subspace $\mathbf{R}\mathcal{S}\subset\mathbb{R}^k$,  with corresponding orthogonal projectors $P_{\mathbf{R}\mathcal{S}}$ and $P_{\mathbf{R}\mathcal{S}}^{\perp}$ in $\mathbb{R}^k$.}

Define the projection residual as
\begin{equation}
  \begin{aligned}
    \mathbf{e}_{\mathrm{proj}}
    &: =\;\underbrace{\mathbf{R}\mathbf{x}}_{\text{projected sample}}
    -\;\underbrace{P_{\mathbf{R}\mathcal{S}}\, \mathbf{R}\mathbf{x}}_{\text{projection onto }\mathbf{R}\mathcal{S}} \\ 
    &=\;P_{\mathbf{R}\mathcal{S}}^{\perp}\mathbf{R}\mathbf{x}
    =\;P_{\mathbf{R}\mathcal{S}}^{\perp}\mathbf{R}\mathbf{x}_{\perp}.
  \end{aligned}
  \label{eq: residual-def}
\end{equation}

From \eqref{eq: residual-def},  we observe: 
\begin{itemize}
  \item \textbf{Unique Error Source: } Only the component of $\mathbf{x}$ orthogonal to $\mathcal{S}$,  namely $\mathbf{x}_{\perp}$,  contributes to the projection residual.
  \item \textbf{Geometric Interpretation: } The vector $\mathbf{e}_{\mathrm{proj}}$ is exactly the orthogonal projection of $\mathbf{R}\, \mathbf{x}_{\perp}$ onto the orthogonal complement of $\mathbf{R}\mathcal{S}$ in $\mathbb{R}^{k}$. Its dimension is $k - p$,  analogous to the classical least‐squares residual lying in $\mathbb{R}^{n-p}$ \cite{CrossValidated2021}.
\end{itemize}

\subsection{Sin $\Theta$ Angle Leakage Bound}

\textcolor{black}{
To quantify the subspace rotation error induced by SRP,  we invoke the Davis–Kahan (Wedin) $\sin\Theta$ theorem \cite{Yu2015, Hsu2016}. Specifically,  if a small perturbation is applied to a subspace,  then the largest principal angle between the original and perturbed subspaces is bounded in proportion to the operator norm of the perturbation. Under the $\varepsilon$-subspace embedding assumption (i.e.,  $\|\mathbf{R}^{\top}\mathbf{R} - \mathbf{I}\|_{2} \le \varepsilon$),  the Davis–Kahan $\sin\Theta$ theorem implies}
\begin{equation}
  \bigl\|\sin\Theta\bigl(\mathcal{S}, \, \mathbf{R}\mathcal{S}\bigr)\bigr\|_{2}
  \;=\;\bigl\|P_{\mathcal{S}} - P_{\mathbf{R}\mathcal{S}}\bigr\|_{2}
  \;\le\;\varepsilon.
  \label{eq: sin-theta}
\end{equation}

Consequently,  for any $\mathbf{x}_{\perp}\perp \mathcal{S}$,  one obtains

\begin{align}
  \|P_{\mathbf{R}\mathcal{S}}\, \mathbf{R}\, \mathbf{x}_{\perp}\|
  &\le \sqrt{2}\, \varepsilon\, \|\mathbf{R}\, \mathbf{x}_{\perp}\|
    \quad\text{(by $\sin\Theta$)} \notag\\
  &\le \sqrt{2}\, (1+\varepsilon)\, \varepsilon\, \|\mathbf{x}_{\perp}\|
    \quad\text{(by $\varepsilon$-embedding)} \notag\\
  &= C_{1}\, \varepsilon\, \|\mathbf{x}_{\perp}\|.
  \label{eq: angle-leakage}
\end{align}

\textcolor{black}{
where $C_{1} : = \sqrt{2}\, (1 + \varepsilon)\approx 2.9$ for $\varepsilon\le 0.3$. Inequality \eqref{eq: angle-leakage} gives an explicit upper bound on the “angle leakage” due to SRP; in subsequent derivations,  we uniformly denote this constant by $C_{1}$.}

\subsection{Sparse Hanson–Wright Concentration Inequality}

\textcolor{black}{
Beyond the subspace rotation error quantified by \eqref{eq: angle-leakage},  one must ensure that the $\ell_{2}$-norm $\bigl\|\mathbf{R}\, \mathbf{x}_{\perp}\bigr\|$ remains tightly concentrated around $\|\mathbf{x}_{\perp}\|$ in a single SRP instance. To this end,  we employ a \emph{sparse Hanson–Wright inequality},  which generalizes the classical Hanson–Wright bound to the setting of $\{0, \pm1\}$-valued matrices with density $1 - s$ \cite{Zhong2024, RudelsonVershynin2013}. In particular,  for any fixed vector $\mathbf{x}_{\perp}$,  if $\mathbf{R}$ is drawn according to the sparse random model of density $1 - s$,  then}
\begin{equation}
  \Pr\Bigl[\bigl|\|\mathbf{R}\, \mathbf{x}_{\perp}\|^{2} 
  \;-\; \|\mathbf{x}_{\perp}\|^{2}\bigr|
  \;>\; \varepsilon\, \|\mathbf{x}_{\perp}\|^{2}\Bigr]
  \;\le\;\frac{\delta}{2}.
  \label{eq: sparse-HW}
\end{equation}
\textcolor{black}{
The proof relies on sub-Gaussian tail bounds for “striped” matrix structures and accounts for the sparsity parameter $1 - s$,  yielding an error bound of the same order as in the dense JL setting \cite{Zhong2024}.}

\subsection{Main Residual Energy Theorem}
\label{subsec: residual-main}

Define the following events: 
\begin{itemize}
  \item $\mathcal{E}_{\mathrm{len}}$:  the length‐preservation event 
    \[
      \Bigl|\|\mathbf{R}\, v\| - \|v\|\Bigr|/\|v\|\;\le\;\varepsilon
      \quad \text{for all }v \in \mathcal{S}\cup\mathcal{S}^{\perp};
    \]
  \item $\mathcal{E}_{\mathrm{ang}}$:  the angle‐leakage event 
    \[
      \bigl\|P_{\mathbf{R}\mathcal{S}}\;\mathbf{R}\, \mathbf{x}_{\perp}\bigr\|
      \;\le\; C_{1}\, \varepsilon\, \bigl\|\mathbf{x}_{\perp}\bigr\|;
    \]
  \item $\mathcal{E}_{\mathrm{hw}}$:  the sparse Hanson–Wright event \eqref{eq: sparse-HW},  i.e.,  
    $\|\mathbf{R}\, \mathbf{x}_{\perp}\|$ is concentrated in $(1 \pm \varepsilon)\, \|\mathbf{x}_{\perp}\|$.
\end{itemize}
\textcolor{black}{
By allocating the total allowable failure probability $\delta$ uniformly across these three key events,  one enforces}
\begin{equation}
  \Pr\bigl[\mathcal{E}_{\mathrm{len}}\bigr] \;, \;\Pr\bigl[\mathcal{E}_{\mathrm{ang}}\bigr] \;, \;\Pr\bigl[\mathcal{E}_{\mathrm{hw}}\bigr]
  \;\ge\; 1 - \frac{\delta}{3}.
\end{equation}
Then,  by the union bound \cite{Feller1971},  the intersection
\[
  \mathcal{E}_{\mathrm{tot}} : = \mathcal{E}_{\mathrm{len}} \;\cap\; \mathcal{E}_{\mathrm{ang}} \;\cap\; \mathcal{E}_{\mathrm{hw}}
\]
\textcolor{black}{
satisfies $\Pr[\mathcal{E}_{\mathrm{tot}}]\ge 1 - \delta$. The reason one can require each sub-event to have failure probability at most $\delta/3$ is that,  in what follows,  one chooses the projection dimension $k$ sufficiently large so that each of the three conditions—$\varepsilon$-subspace embedding,  Davis–Kahan angle bound,  and sparse Hanson–Wright concentration—fails with probability at most $\delta/3$. Hence the overall failure probability does not exceed $\delta$.}

Under event $\mathcal{E}_{\mathrm{tot}}$,  let $C_{2} : = 1 + C_{1}$. Then,  for any decomposition $\mathbf{x} = \mathbf{x}_{\mathcal{S}} + \mathbf{x}_{\perp}$,  it holds that
\begin{gather}
  \bigl(1 - C_{2}\, \varepsilon\bigr)\, \|\mathbf{x}_{\perp}\|
  \;\le\; \|\mathbf{e}_{\mathrm{proj}}\|
  \;\le\;\bigl(1 + C_{2}\, \varepsilon\bigr)\, \|\mathbf{x}_{\perp}\|, 
  \label{eq: residual-bound-1} \\
  \bigl|\|\mathbf{e}_{\mathrm{proj}}\|^{2} - \|\mathbf{x}_{\perp}\|^{2}\bigr|
  \;\le\;2\, C_{2}\, \varepsilon\, \|\mathbf{x}_{\perp}\|^{2}.
  \label{eq: residual-bound-2}
\end{gather}

\begin{figure*}[t]
  \centering 
  \includegraphics[width=1.01\textwidth]{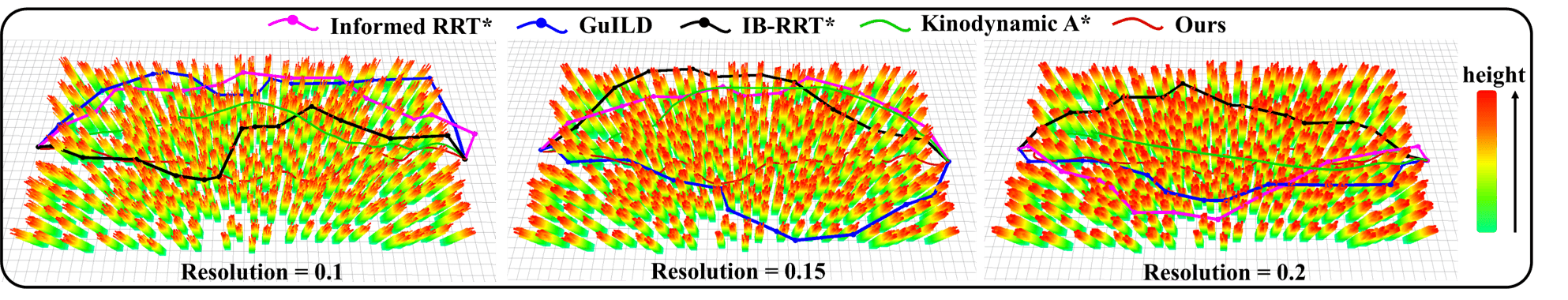}
  \caption{\textcolor{black}{This figure showcases the performance of our gradient-optimized front-end search in a dense,  cluttered map. Leveraging the occupancy field gradient,  our method proactively guides trajectories into \textcolor{black}{safer areas}. This effectively avoids the path quality degradation of standard Kinodynamic A* caused by upward planning in dense areas,  and significantly outperforms the lengthy,  randomized paths of RRT*-based methods. Ultimately,  our approach generates safer,  smoother,  and more efficient paths across various resolutions.}}
  \label{front_end_multi_resolution_contrast}
\end{figure*}

\textcolor{black}{
The \emph{Residual Energy Preservation Theorem} above combines the $\varepsilon$-subspace embedding property with classical linear-model error bounds to fill a gap in random mapping theory:  it gives a quantitative analysis of how the energy of the component orthogonal to $\mathcal{S}$ is preserved after projection. Specifically,  when the projection dimension satisfies \eqref{eq: ose-dim},  inequalities \eqref{eq: residual-bound-1}–\eqref{eq: residual-bound-2} hold with $C_{2} = 1 + \sqrt{2}\, (1 + \varepsilon)\approx 3.9$ (for $\varepsilon \le 0.3$). This result provides clear guidance for choosing dimensionality‐reduction parameters:  one can ensure that the projected residual remains within an $O(\varepsilon)$ factor of the original residual energy,  and one can reduce the failure probability $\delta$ by increasing $k$ appropriately. In large‐scale sparse feature settings,  SRP not only reduces the matrix–vector multiplication cost from $O(d\, k)$ to $O(s\, k)$,  but also significantly accelerates computation while preserving approximation accuracy,  thereby guaranteeing model stability and generalization after projection.}

\textcolor{black}{The effectiveness of the RMRP framework is validated by the following results. Both our qualitative analysis (\textcolor{black}{Fig.~\ref{classify},  ~\ref{regression}}) and quantitative benchmarks (\textcolor{black}{Fig.~\ref{data}}) confirm that RMRP maintains high accuracy in classification and regression tasks while significantly reducing computation time.}









\section{Unified Linear Parametric Map}
Occupancy grid map can distinguish between occupied,  free,  and unknown regions of the environment,  which is used for path searching in the front-end. Meanwhile,  ESDF maps can provide Euclidean distance and gradient values for trajectory optimization in the back-end. 

The core challenge in occupancy mapping involves estimating spatial occupancy states. Traditional Bayesian methods require storing substantial prediction data for occupancy evaluation. Generating such discrete maps and accessing occupancy states for path searching entails significant storage and computational demands,  especially in large-scale environments.

In ESDF mapping,  calculating and storing Euclidean distances from each voxel center to the nearest obstacle is essential. Distance and gradient values at arbitrary points are typically derived through interpolation at appropriate resolutions,  yet balancing computational efficiency with accuracy remains challenging.

\begin{figure}[h]
	\centering 
	\includegraphics[width=0.49\textwidth]{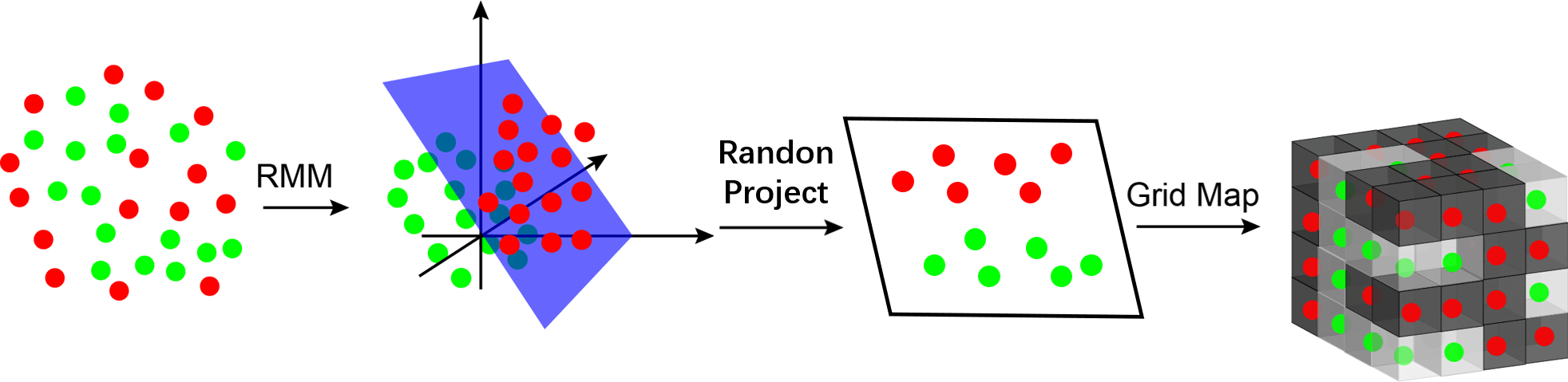   }
	\caption{\textcolor{black}{RMRP Mapping Pipeline for Dimensionality Reduction and Geometric Preservation:  The point cloud is initially mapped to a high-dimensional space to achieve linear separability. Subsequently,  efficient dimensionality reduction is performed using a sparse random projection that preserves key geometric structures. The resulting occupied (red) and free  (green) voxels remain distinctly classified post-reduction.}
 } 
	\label{method}%
\end{figure}

To overcome these issues,  we propose a lightweight linear parametric model for occupancy mapping and ESDF mapping.
Firstly,  real-time autonomous flight in large-scale environments,  UAVs acquire irregular point cloud datasets $\bm{D}=\left\{x_i,  t_i\right\}_{i=1}^L$. Data in the original low-dimensional space exhibits irregular separability.  We employ RMRP to map these disordered points into a high-dimensional space,  which are approximately linearly separable. \textcolor{black}{This complete mapping pipeline is visually summarized in \textcolor{black}{Fig.~\ref{method}}.} This transformation enables environment representation via a simplified linear model,  substantially reducing computational costs for subsequent occupancy and distance estimations. It is worth noting that RMM requires a higher dimension to improve accuracy,  but it also increases computational burden. We use RMRP to reduce dimensions to improve computing efficiency while maintaining accuracy. 

In an occupancy grid map,  a discrete target vector t represents the occupancy state (0 for free and 1 for occupied). We use the RMRP to transform the occupancy mapping into a classification task,  thereby learning a linear parametric model as follows: 
\begin{equation}
  f = f(x,  \eta) =  \bm{\eta}^T \bm{R} \bm{g}( \bm{W} \bm{x}) + b = \bm{\eta}^T S' + b= \bm{\eta}^T S' 
  \label{grid_MAP_MDOEL}
\end{equation}

In an ESDF map,  a continuous target vector t denotes the Euclidean distance to the nearest obstacle. By RMRP,  the ESDF mapping is treated as a regression problem,  we can obtain a linear parametric model to establish the relationship between position and Euclidean distance. The ESDF model is as follows: 
\begin{equation}
 D = \bm{\beta}^T \bm{R} \bm{g}( \bm{W} \bm{x}) 
 \label{D_esdf}
\end{equation}

\subsection{\textcolor{black}{Predictive Scene Completion}}

\textcolor{black}{To address the poor extrapolation capability inherent in function approximation-based mapping,  we introduce a supervised learning paradigm,  conceptually illustrated in \textcolor{black}{Fig. 9}. This approach involves an offline training phase (\textcolor{black}{Algorithm ~\ref{alg: offlinetraining}}) using complete 3D scenes,  which compels the RMRP model to learn the latent geometric priors between visible structures and their occluded regions. During online deployment,  the model leverages these learned priors to perform robust scene completion by inferring obstacle states in blind spots,  a process managed by \textcolor{black}{Algorithm ~\ref{alg: OnlineProcess}}. This predictive capability enables proactive and safer trajectory generation,  with its effectiveness validated through both qualitative completion results (\textcolor{black}{Fig. 11,  12}) and significant quantitative safety improvements (\textcolor{black}{Fig. 14}).}

\begin{figure}[t]
	\centering 
	\includegraphics[width=0.48\textwidth]{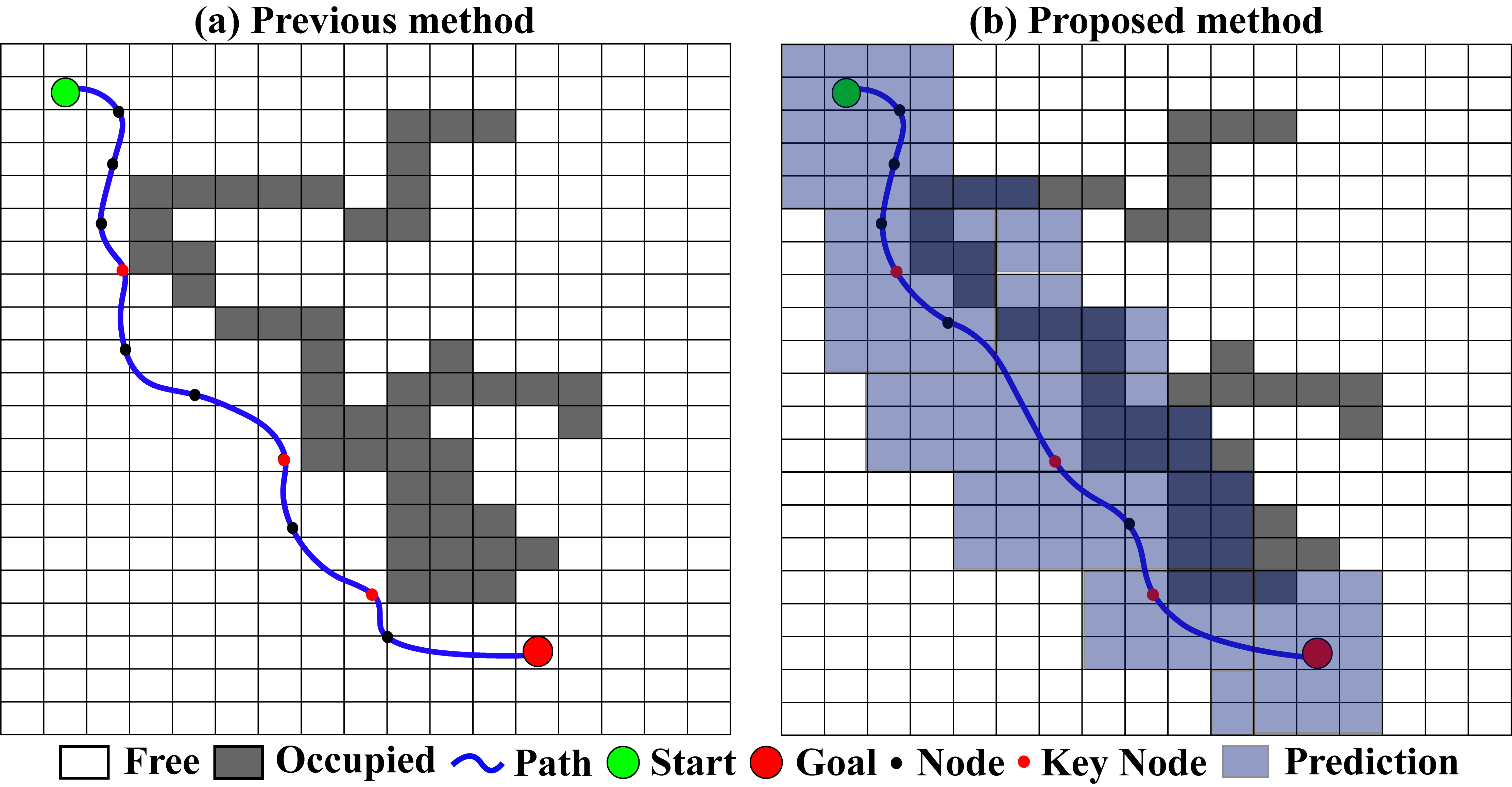}
	\caption{\textcolor{black}{By incorporating the occupancy field gradient generated by the RMRP model for the prediction region,  our optimizer actively pushes the trajectory away from regions with abrupt occupancy changes (i.e.,  obstacle boundaries). This effectively mitigates unnecessary cornering behavior near key nodes in kinematically constrained paths,  resulting in significant global trajectory smoothing while ensuring safety.}}
	\label{pathPlanning}
\end{figure}

\begin{figure}[t]
	\centering 
	\includegraphics[width=0.48\textwidth]{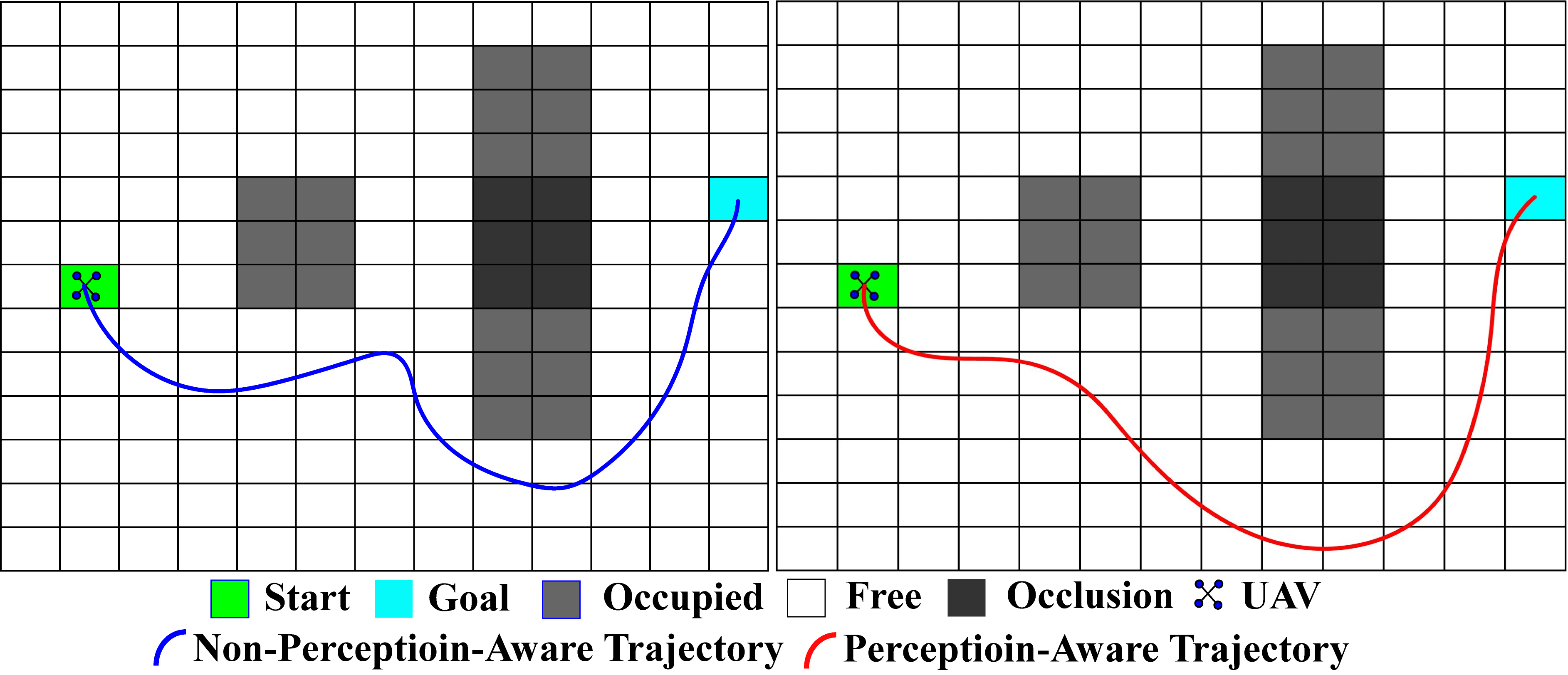}
	\caption{\textcolor{black}{Traditional methods (right) can only plan based on current observations. Our method (left),  however,  employs the RMRP model for predictive completion of sensor blind spots. This allows the planner to acquire more complete environmental information in advance,  enabling the generation of globally optimal paths that proactively avoid risks.}}
    \vspace{-15pt} 
    \label{fig: introduction_1}
\end{figure}

\begin{figure}[t]
	\centering 
	\includegraphics[width=0.48\textwidth]{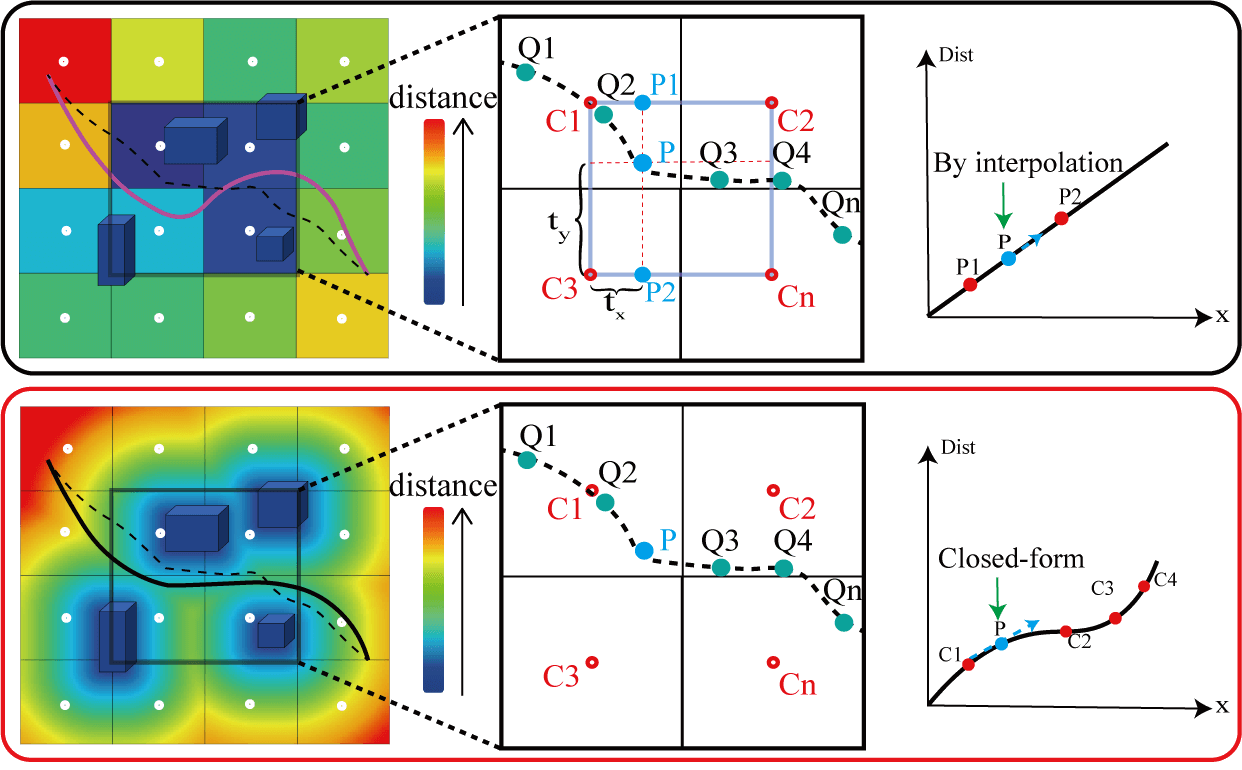}
	\caption{\textcolor{black}{Conventional methods (top) rely on discrete maps and interpolated distance estimates,  resulting in limited accuracy. Our method (bottom) obtains a closed-form ESDF solution through a parametric model,  enabling the analytical computation of precise distances and guiding the optimizer to generate smoother and safer trajectories.}}
    \vspace{-15pt} 
	\label{backend_ESDF_closed}
\end{figure}

    \vspace{-10pt} 

\begin{algorithm}
\caption{\textcolor{black}{Offline Model Training (RMRP)}}\label{alg: offlinetraining}
\KwIn{Historical dataset $(X,  T)$; RM dim. $M$; SRP dim. $k$ $(k < M)$; reg. param. $\alpha$}
\KwOut{Pretrained model $(\eta,  W,  b,  g,  R)$}

$W,  b \leftarrow$ \textbf{RandInit}$(M,  \dim(X))$\;

$R \leftarrow$ \textbf{SparseProj}$(k,  M)$\;

$S \leftarrow g(WX + b)$\;

$S' \leftarrow R S$\;

$\eta \leftarrow$ \textbf{RidgeSolve}$(S',  T,  \alpha)$\;

\Return $(\eta,  W,  b,  g,  R)$\;
\end{algorithm}

\textcolor{black}{\subsection{Trajectory Refinement for Frontend Path Search}}

\textcolor{black}{In robot motion planning,  frontend search-based planners,  such as Kinodynamic A*,  are powerful tools for finding dynamically feasible trajectories. Such algorithms can efficiently find an initial trajectory,  $P_{init}$,  that connects the start and goal configurations in complex environments. However,  owing to their nature of searching over a discretized state space,  the resulting trajectories often exhibit suboptimal smoothness and insufficient safety margins,  manifesting as paths with unnecessary maneuvers or those that pass in close proximity to obstacle boundaries.}

\textcolor{black}{To significantly enhance the quality of the initial trajectory,  we introduce a subsequent trajectory refinement stage. This process is designed to optimize the initial feasible solution provided by Kinodynamic A*,  making it smoother and safer while preserving its fundamental reachability. The core of this refinement process lies in leveraging the continuous and differentiable environment representation provided by RMRP to formulate an optimization problem.}

\textcolor{black}{The objective of the refinement is to adjust the initial sequence of waypoints,  $P_{init} = \{x_0,  x_1,  ...,  x_p\}$,  by minimizing a specially designed cost function $J(P)$: }

\begin{equation}
  P^* \;=\; \underset{P}{\arg\min}\; J(P)
  \label{eq: opt_waypoints}
\end{equation}

\textcolor{black}{The cost function $J(P)$ is specifically formulated as: }

\begin{equation}
  J(P)
  = \lambda_g \sum_{i=1}^{p-1} \left\| \frac{\partial y}{\partial x_i} \right\|^2
    + \lambda_d \sum_{i=1}^{p-1} \bigl\lVert x_i - x_{i, \mathrm{init}}\bigr\rVert^2
  \label{eq: cost_function}
\end{equation}

\textcolor{black}{The first term in this cost function is key to the refinement. By minimizing the sum of squared norms of the occupancy change rate (i.e.,  the occupancy field gradient) at each waypoint,  it acts as a repulsive force,  pushing the trajectory away from obstacle boundaries where the occupancy probability changes sharply. This mechanism directly addresses the issue of proximity to obstacles found in the initial path and naturally guides the trajectory towards a smoother profile. The resulting global trajectory smoothing effect is clearly illustrated in \textcolor{black}{Fig. 8},  where sharp cornering is effectively mitigated. The occupancy field gradient is computed as follows: }

\begin{equation}
  \frac{\partial y}{\partial x}
  = \frac{\partial\bigl(\eta^{T}Rg(Wx + b)\bigr)}{\partial x}
  = \eta^{T} \,  J_{g}(x_{1},  x_{2},  \dots,  x_{N})
  \label{eq: occupancy_gradient}
\end{equation}

\textcolor{black}{where $J_g(\cdot)$ is the Jacobian matrix,  which is uniquely determined for a trained model. The second term,  meanwhile,  acts as a regularization term,  ensuring that the refinement process does not drastically alter the global structure of the initial path,  thereby preserving the feasible corridor found by the frontend search. The entire optimization problem can be solved using standard optimization techniques. Furthermore,  in practical implementations,  it is crucial to apply scale normalization to the different cost terms to ensure balanced optimization. This entire refinement process is summarized in \textcolor{black}{Algorithm~\ref{alg: refinement_frontend}}.}

\begin{algorithm}
\caption{\textcolor{black}{Process Next Online Event with RMRP}}\label{alg: OnlineProcess}
\KwIn{Model $M_k = (\eta_k,  W,  b,  g,  R)$; map $\mathcal{M}_k$; event $E$; AdamW params $(\gamma,  \lambda,  \eta_1,  \eta_2,  \epsilon)$; threshold $\tau$}
\KwOut{Updated $M_{k+1}$ and $\mathcal{M}_{k+1}$}

\If{$E = (x_k,  t_k)$ (data)}{
    \tcc{\textcolor{blue}{\textbf{Incremental Learning}}}
    $(\eta_k,  W,  b,  g,  R) \leftarrow M_k$\;
    $s_k \leftarrow R \cdot g(W x_k + b)$\;
    $\text{grad}_k \leftarrow -(t_k - \eta_k^\top s_k) \cdot s_k$\;

    \tcc{\textcolor{blue}{\textbf{Online Optimization}}}
    $\eta_{k+1} \leftarrow \textbf{AdamW}(\eta_k,  \text{grad}_k,  \gamma,  \lambda,  \eta_1,  \eta_2,  \epsilon)$

    $M_{k+1} \leftarrow (\eta_{k+1},  W,  b,  g,  R)$\;
    $\mathcal{M}_{k+1} \leftarrow \mathcal{M}_k$\;
    \Return $(M_{k+1},  \mathcal{M}_{k+1})$\;
}
\ElseIf{$E = X^* = \{x^*_1,  \dots,  x^*_p\}$ (query)}{
    \tcc{\textcolor{blue}{\textbf{Prediction Phase}}}
    $Y^* \leftarrow [\ ]$\;
    \For{each $x^*_j \in X^*$}{
        $s^*_j \leftarrow R \cdot g(W x^*_j + b)$\;
        $y^*_j \leftarrow \eta_k^\top s^*_j$\;
        Append $y^*_j$ to $Y^*$\;
    }

    \tcc{\textcolor{blue}{\textbf{Obstacle Completion}}}
    $\mathcal{M}_{k+1} \leftarrow \mathcal{M}_k$\;
    \For{each $(x^*_j,  y^*_j) \in \text{zip}(X^*,  Y^*)$}{
        \If{$y^*_j > \tau$}{
            $\mathcal{M}_{k+1}(x^*_j) \leftarrow 1$ \tcp*{Occupied}
        }
    }

    $M_{k+1} \leftarrow M_k$ \tcp*{Model unchanged}
    \Return $(M_{k+1},  \mathcal{M}_{k+1})$\;
}
\end{algorithm}

\subsection{\textcolor{black}{Trajectory Optimization for UAVs}}

In this paper,  We employ B-spline  \cite{zhou2019robust}  for gradient-based trajectory optimization. For a \textcolor{black}{$p_b$-degree} B-spline trajectory defined by \textcolor{black}{$N+1$} control points $\left\{ Q_0,  Q_1,  ...,  Q_N \right\}$,  the overall cost function is defined as:

\vspace{-5pt}
\begin{equation}
\begin{aligned}
f_{total} &= \lambda_s f_s + \lambda_c f_c + \lambda_d (f_v + f_a) \\
&= \lambda_s \sum_{i=p_b-1}^{N-p_b+1} \|\mathbf{Q}_{i+1} - 2 \mathbf{Q}_i + \mathbf{Q}_{i-1}\|^2
\end{aligned}
\end{equation}
\vspace{-15pt}
\begin{align*}
&\hspace{5.6em}+ \lambda_c \sum_{i=p_b}^{N-p_b} \mathbf{F}(d(\mathbf{q}_i),  d_{\min}) \\
&\hspace{-1em}+ \lambda_d \sum_{\{x, y, z\}} \left\{ \sum_{i=p_b-1}^{N-p_b} \mathbf{F}(v_{\max}^2,  \tilde{q}_{i, \mu}^2) + \sum_{i=p_b-2}^{N-p_b} \mathbf{F}(a_{\max}^2,  \tilde{q}_{i, \mu}^2) \right\}
\end{align*}
where $f_s$ and $f_c$ are smoothness and collision cost. $f_v$ and $f_A$ are dynamic limits on velocity and acceleration. Coefficients  $\lambda_s$,  $\lambda_c$ and $\lambda_d$ mediate trade-offs between trajectory smoothness,  obstacle safety,  and dynamic feasibility.

Here $F()$ is a penalty function for general variables as follows: 
\begin{equation}
F(x, y) =
\begin{cases} 
(x - y)^2,  & x  \le y \\  
0,  & x >y
\end{cases}
\end{equation}
Collision Penalty:  The trajectory is optimized on the local ESDF map to avoid obstacles safely.

\begin{equation}
\left.F_c(\mathbf{D(\mathbf{Q}_i)})=\left\{\begin{matrix}(\mathbf{D(\mathbf{Q}_i)}-S_{f})^2&\mathbf{D(\mathbf{Q}_i)}\leq S_{f}\\0&\mathbf{D(\mathbf{Q}_i)}>S_{f}\end{matrix}\right.\right.
\end{equation}

The Euclidean distance $D(Q_i)$,  derived from \eqref{D_esdf},  must exceed the minimum clearance $S_f$ required between obstacles and the trajectory. Unlike conventional ESDF methods dependent on interpolation,  our approach utilizes a closed-form ESDF map to acquire distance information,  \textcolor{black}{a conceptual comparison that is visually detailed in \textcolor{black}{Fig. 10}}. And the gradients are computed analytically by computing the
derivative of $F_c$ with respect to $Q_i$.

\begin{figure}[h]
	\centering 
	\includegraphics[width=0.49\textwidth]{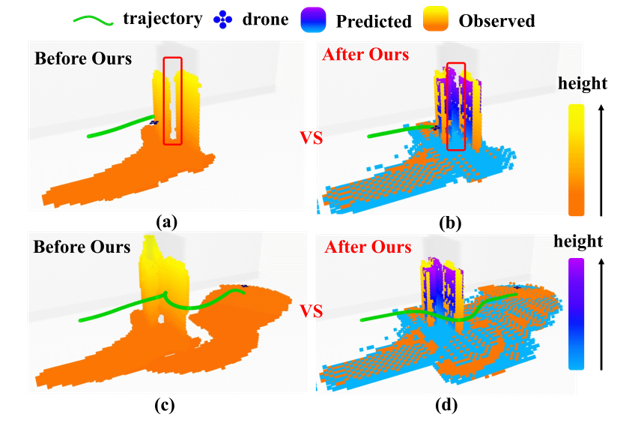   }
	\caption{\textcolor{black}{Scenario 1:  Utilizing the RMRP model,  Ours proactively completes the sensor blind spot behind the obstacle (blue point cloud) based on the observed data (yellow point cloud). This enables the generation of a safer,  globally evasive trajectory,  avoiding reactive emergency obstacle avoidance.}
 } 
	\label{drone_scene1_predict}%
\end{figure}

\begin{figure}[h]
	\centering 
	\includegraphics[width=0.49\textwidth]{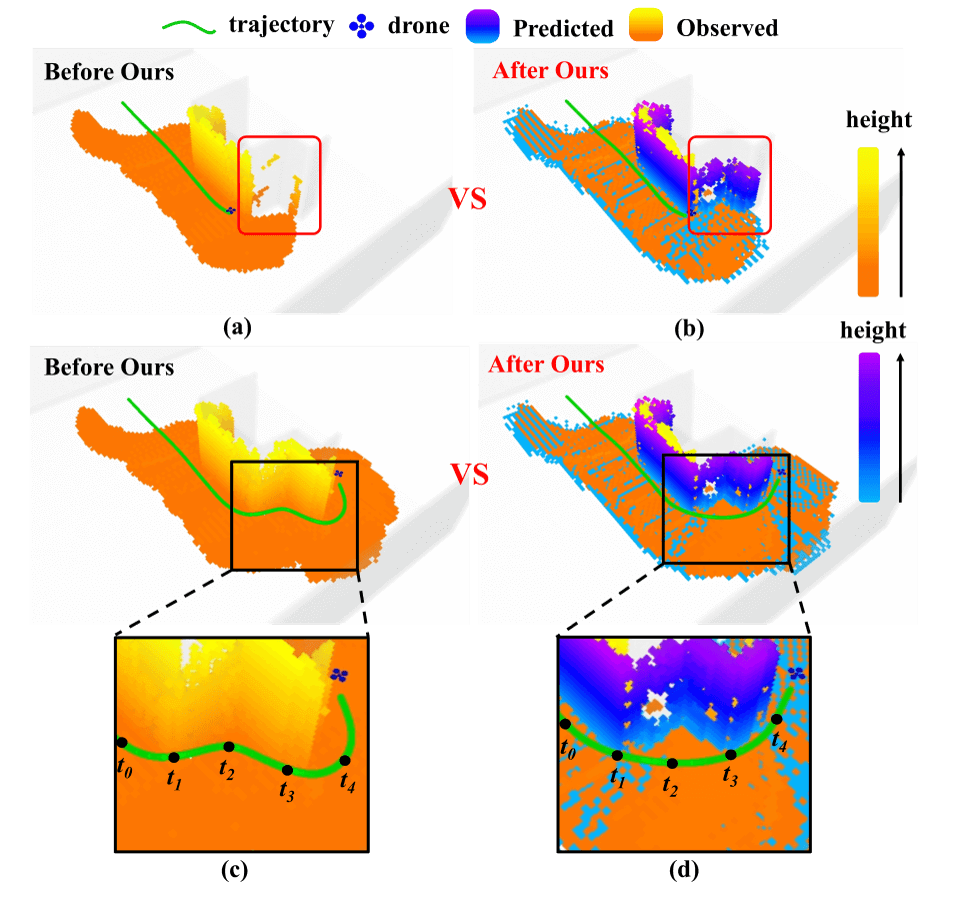   }
	\caption{\textcolor{black}{Scenario 2:  Completion and Planning in Lateral Blind Spots at Corners. The left image (Before) shows a risky trajectory closely adhering to the inside of the corner in the absence of predictive capabilities. In the right image (After),  our RMRP model completes the corner's blind spot and plans a proactive avoidance path. As highlighted in the local magnification,  the increased safety margin between the new trajectory and the corner significantly enhances safety and robustness during high-speed cornering.}
 } 
	\label{drone_scene2_predict}%
\end{figure}

\begin{algorithm}
\caption{\textcolor{black}{Refinement of Frontend Path}}\label{alg: refinement_frontend}
\KwIn{RMRP model $(\eta,  W,  b,  g)$; start/goal $x_s$,  $x_g$; weights $\lambda_g$,  $\lambda_d$; iterations $N$; step size $\alpha$}
\KwOut{Refined trajectory $P^*$}

$P_0 \leftarrow$ \textbf{KinodynamicAStar}$(x_s,  x_g,  \text{RMRP})$\;
$P^* \leftarrow P_0$\;

\For{$k \leftarrow 1$ \KwTo $N$}{
    $\nabla J \leftarrow$ \textbf{Grad}$(P^*,  P_0,  \text{RMRP},  \lambda_g,  \lambda_d)$\;
    $P^* \leftarrow P^* - \alpha \cdot \nabla J$\;
}
\Return $P^*$
\end{algorithm}

\begin{figure}[h]
	\centering 
	\includegraphics[width=0.49\textwidth]{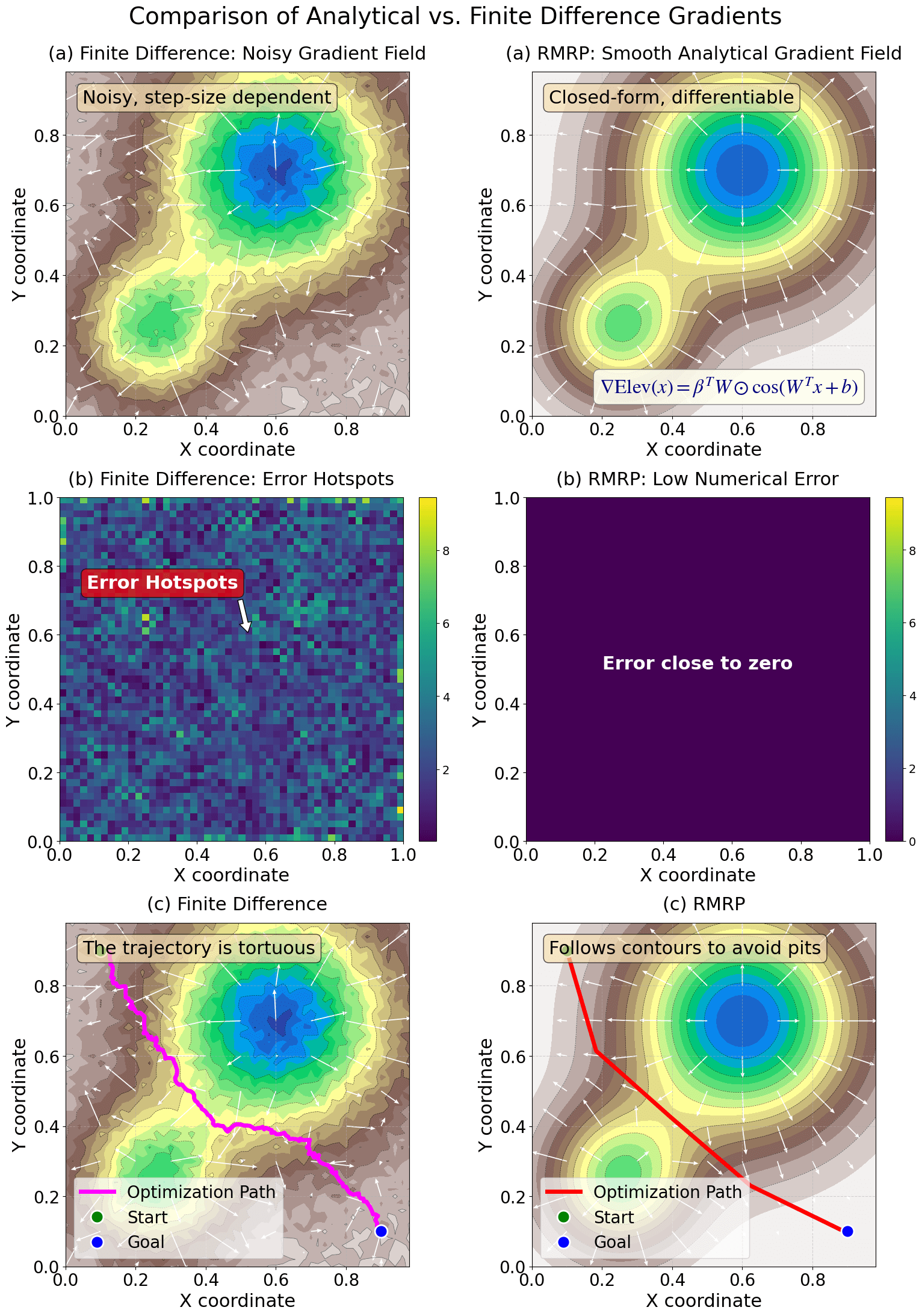   }
	\caption{
\textcolor{black}{The figure clearly demonstrates that RMRP generates a smooth,  accurate gradient field with near-zero numerical error. In contrast,  the finite difference method produces a noisy gradient field with significant error hotspots. This fundamental difference in gradient quality directly impacts the final path quality and efficiency:  the RMRP-based optimizer plans an efficient,  smooth path,  while the noisy gradient-based optimizer,  although reaching the goal,  generates a tortuous and inefficient path.}
 } 
	\label{analytical_vs_finite_difference_gradient}%
\end{figure}

\begin{figure}[h]
	\centering 
	\includegraphics[width=0.49\textwidth]{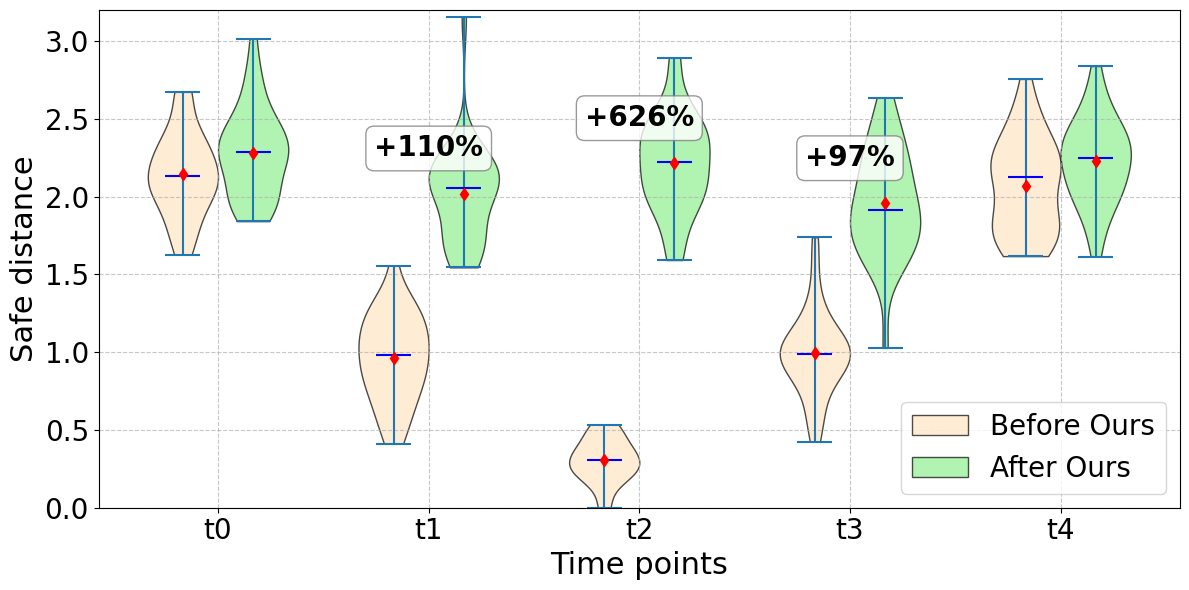   }
	\caption{
    \textcolor{black}{The plot compares the safe distance between the drone and the blind spot center at five key instants,  both without (beige) and with (light green) our method. The results clearly indicate that our method maintains a larger safety margin throughout the process,  with the median distance increasing by over 600\% at the most critical $t_2$ instant. This strongly validates that online environmental completion can significantly enhance the flight safety of drones in unknown environments.}} 
	\label{violinplot}%
\end{figure}

\begin{figure*}[t]
  \centering 
  \includegraphics[width=1.02\textwidth]{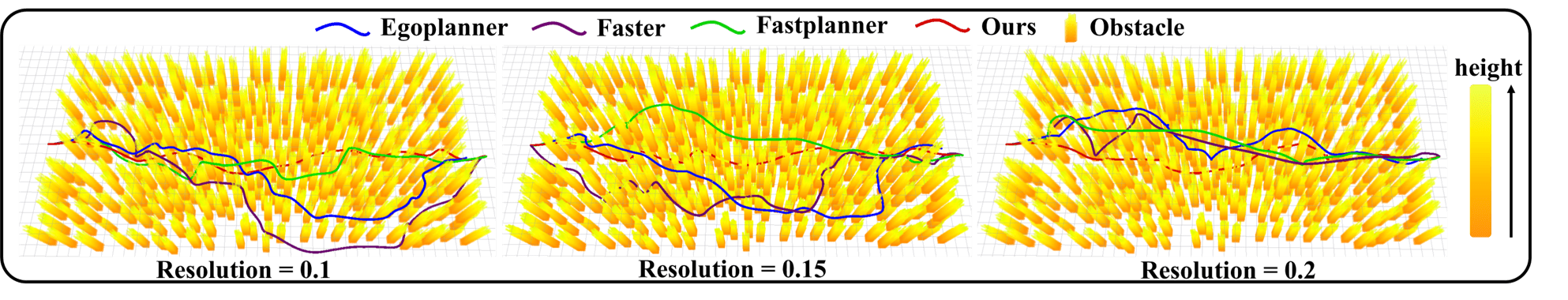}
  \caption{\textcolor{black}{This figure illustrates the sensitivity of different planners' trajectory optimization to map resolution in a dense obstacle environment. Traditional planners (e.g.,  Fast-Planner,  Faster),  which rely on discrete ESDF maps,  and EGO-Planner,  which bypasses ESDF but still requires an environment map for collision detection,  exhibit trajectory quality variations with changing resolution. Our method,  based on a continuous parametric model,  generates robust and high-quality trajectories that are invariant to resolution.}}
  \label{back_end_multi_resolution_contrast}
\end{figure*}

\section{\textcolor{black}{Terrain-Aware Trajectory Optimization}}

\textcolor{black}{For UGVs operating in complex and unstructured environments,  the capability for safe and efficient navigation is paramount. A central challenge in achieving this lies in the accurate evaluation and quantification of terrain traversability. Traditional methods based on discrete grids or elevation maps are difficult to employ directly in modern gradient-based trajectory optimization frameworks due to their discontinuous representations. Meanwhile,  models based on methods such as Gaussian Processes (GPs),  while offering continuity,  are limited by their substantial computational overhead,  which restricts their application in real-time,  large-scale scenarios.}

\textcolor{black}{To overcome these limitations,  we leverage the RMRP framework proposed in the preceding sections. This framework not only generates a lightweight and continuous terrain model but,  crucially,  this model is also fully differentiable. This section details how this superior characteristic of the RMRP model is exploited to construct a highly efficient terrain-aware trajectory optimizer by deriving the closed-form gradient of the terrain elevation. The superior quality and accuracy of this analytical gradient field,  compared to numerical methods,  are visually demonstrated in \textcolor{black}{Fig. 13}.}

\begin{figure}[h]
	\centering 
	\includegraphics[width=0.5\textwidth]{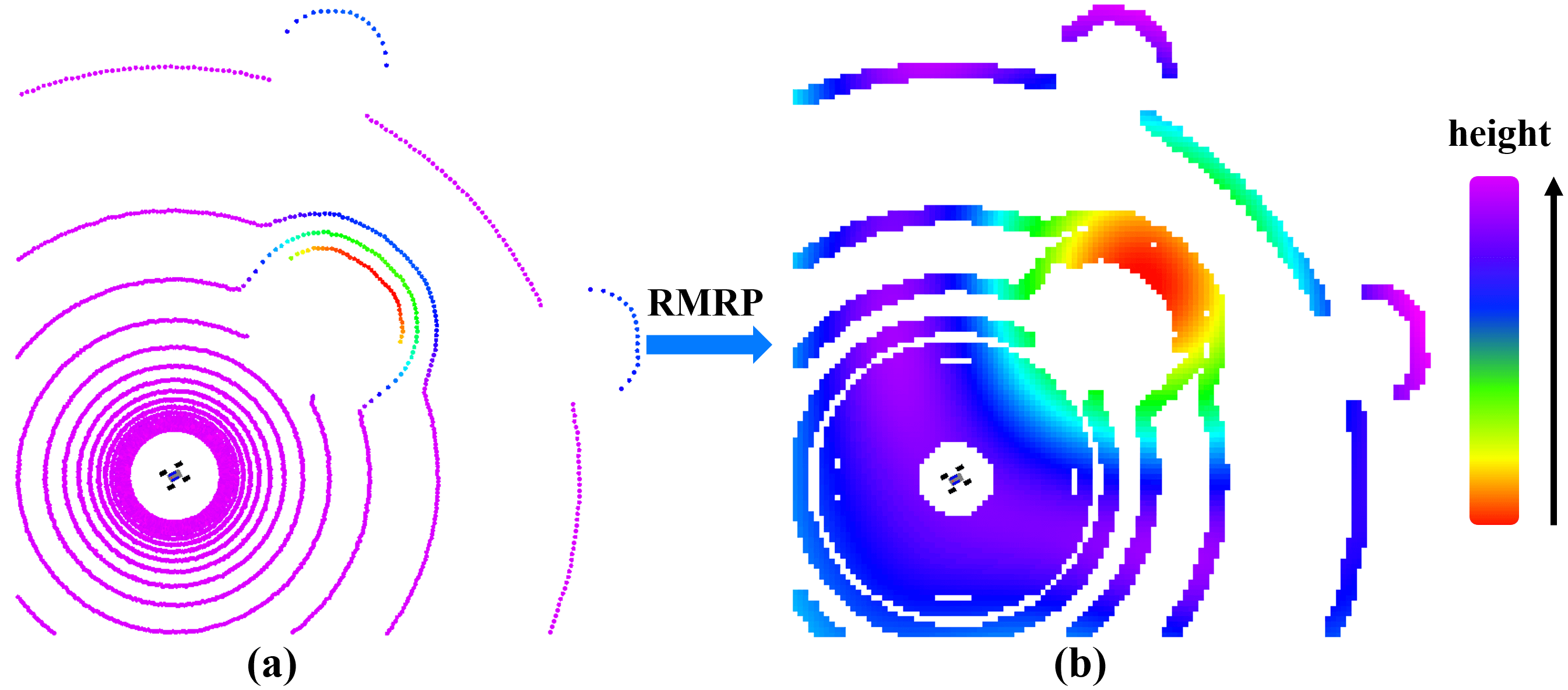  }
	\caption{\textcolor{black}{(a) Raw LiDAR Point Cloud Input. (b) Continuous and Differentiable Terrain Model Generated by RMRP.}
    } 
	\label{car_regression}%
\end{figure}

\begin{figure}[h]
	\centering 
	\includegraphics[width=0.5\textwidth]{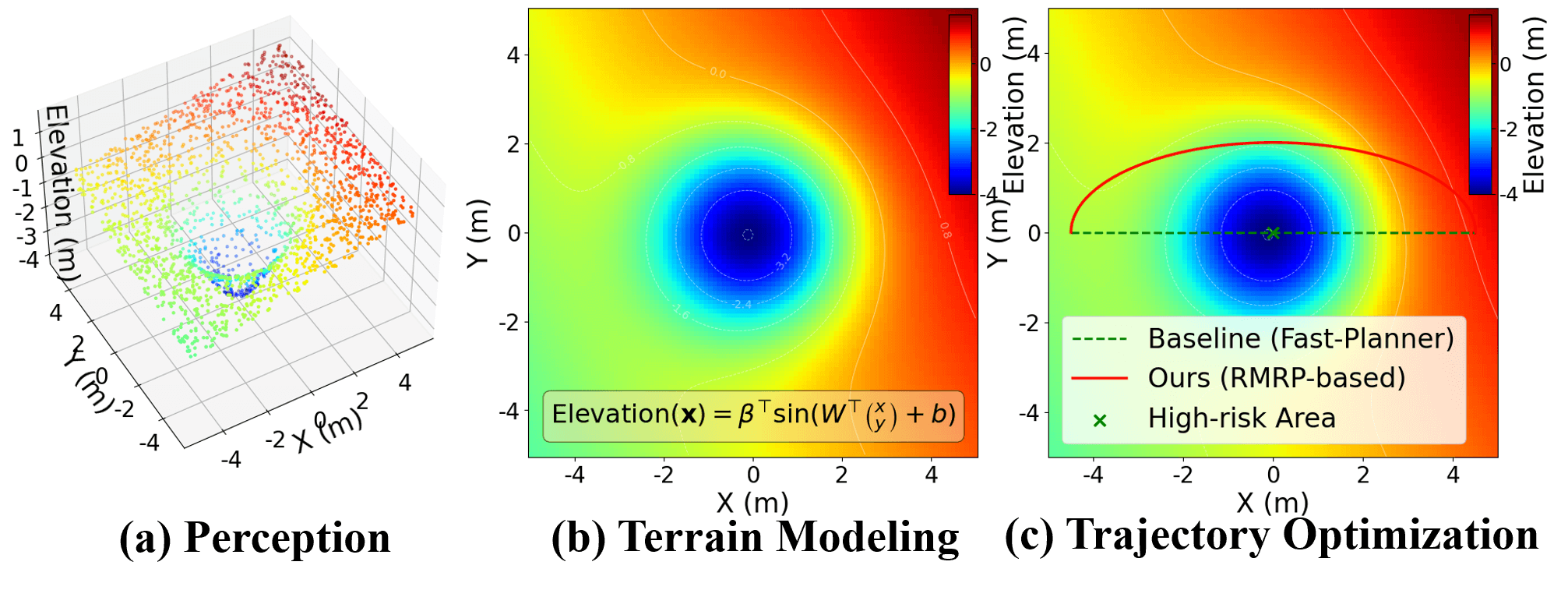  }
	\caption{
    \textcolor{black}{Complete Planning Pipeline Based on Terrain-Aware RMRP Model. (a) Perception:  Raw 3D Point Cloud. (b) Modeling:  Continuous Terrain Map via RMRP. (c) Planning:  Trajectory Optimization Comparison.}} 
	\label{trajectory_optimization_plot}%
\end{figure}

\textcolor{black}{We begin by representing the terrain elevation model as follows: }
\begin{equation}
\mathrm{Elevation}(x) = \beta^T \, \sin\!\bigl(W^T x + b\bigr)
\label{elevation_function}
\end{equation}
where the model parameters are defined as: 
\begin{equation}
x\in\mathbb{R}^2, \; W\in\mathbb{R}^{2\times D}, \; b\in\mathbb{R}^D, \; \beta\in\mathbb{R}^D.
\label{elevation_parameters}
\end{equation}
\textcolor{black}{Here,  $x$ represents the 2D ground coordinates,  while $\beta,  W, $ and $b$ are the model parameters learned via the RMRP method.}

\subsection{Analytical Derivation of the Terrain Elevation Gradient}

\textcolor{black}{The local gradient of the terrain,  or its slope,  is a core physical quantity for evaluating traversability. Our objective is to compute the gradient of the elevation function,  $\nabla \mathrm{Elevation}(x)$,  with respect to the 2D coordinates $x$.}

\textcolor{black}{First,  we expand the elevation function (\ref{elevation_function}) into its summation form: }
\begin{equation}
\mathrm{Elevation}(x) = \sum_{i=1}^D \beta_i \, \sin\!\bigl(W_i^T x + b_i\bigr)
\label{elevation_expanded}
\end{equation}
\textcolor{black}{where $W_i$ is the $i$-th column of the weight matrix $W$,  with dimension $2\times1$.}

\textcolor{black}{Leveraging the linearity of the gradient operator,  the total gradient can be decomposed into the sum of the individual gradients: }
\begin{equation}
\nabla \mathrm{Elevation}(x) = \nabla_x\left[\sum_{i=1}^D \beta_i \sin \left(W_i^T x+b_i\right)\right]
\label{gradient_step1}
\end{equation}
\begin{equation}
= \sum_{i=1}^D \beta_i \nabla_x\left[\sin \left(W_i^T x+b_i\right)\right]
\label{gradient_step2}
\end{equation}

\textcolor{black}{Next,  we apply the chain rule to each term: }
\begin{equation}
\nabla_x\left[\sin \left(W_i^T x+b_i\right)\right] = \cos \left(W_i^T x+b_i\right) \nabla_x\left(W_i^T x+b_i\right)
\label{gradient_step3}
\end{equation}
\begin{equation}
= \cos \left(W_i^T x+b_i\right) W_i
\label{gradient_step4}
\end{equation}

\textcolor{black}{Substituting (\ref{gradient_step4}) back into (\ref{gradient_step2}),  we obtain the final summation form of the gradient: }
\begin{equation}
\nabla \mathrm{Elevation}(x) = \sum_{i=1}^D \beta_i\, \cos\!\bigl(W_i^T x + b_i\bigr)\;W_i
\label{gradient_final}
\end{equation}

\textcolor{black}{For computational convenience,  we write this in a more compact matrix form: }
\begin{equation}
\nabla \mathrm{Elevation}(x) = W\, \bigl(\beta \odot \cos(W^T x + b)\bigr)
\label{gradient_matrix_form}
\end{equation}
\textcolor{black}{Here,  the $\cos(\cdot)$ function is applied element-wise to the vector,  and $\odot$ denotes the Hadamard (element-wise) product.}

\subsection{Terrain Penalty Term for Trajectory Optimization}

\textcolor{black}{The norm of the terrain gradient directly quantifies the steepness of the terrain,  making it a key component in designing a terrain-aware optimization objective.}
\begin{equation}
\bigl\lVert\nabla \mathrm{Elevation}(x)\bigr\rVert = \bigl\lVert W\, \bigl(\beta \odot \cos(W^T x + b)\bigr)\bigr\rVert
\label{gradient_norm}
\end{equation}
\textcolor{black}{A larger norm value indicates a steeper slope,  which corresponds to higher traversal risk or energy consumption.}

\textcolor{black}{For a ground robot,  a trajectory is typically described by a sequence of 2D control points $\{Q_i\in\mathbb{R}^2\}$,  while its height along the z-axis is determined by the terrain model. We formulate the following trajectory penalty term,  $f_{\text{terrain}}$,  based on the squared norm of the terrain gradient: }
\begin{equation}
f_{\text {terrain }} =\sum_{i=p_b}^{N-p_b} \| \nabla \text { Elevation }\left(Q_i\right) \|^2
\label{terrain_penalty1}
\end{equation}
\begin{equation}
=\sum_{i=p_b}^{N-p_b}\left\|W\left(\beta \odot \cos \left(W^T Q_i+b\right)\right)\right\|^2
\label{terrain_penalty2}
\end{equation}
\textcolor{black}{By minimizing this penalty term,  the optimizer is encouraged to generate smoother and safer paths.}

\subsection{Gradient of the Terrain Penalty Term}

\textcolor{black}{To efficiently use the aforementioned penalty term within a nonlinear optimizer (e.g.,  NLopt),  we also require the gradient of $f_{\text{terrain}}$ with respect to each control point $Q_i$.}

\textcolor{black}{For clarity,  we first define the terrain gradient vector: }
\begin{equation}
\mathrm{g}\left(Q_i\right)=\nabla \text { Elevation }\left(Q_i\right)=W\left(\beta \odot \cos \left(W^T Q_i+b\right)\right)
\label{terrain_gradient}
\end{equation}
\textcolor{black}{and the penalty term at a single control point: }
\begin{equation}
f_{\text {terrain },  i}=\left\|\mathbf{g}\left(Q_i\right)\right\|^2
\label{terrain_penalty_i}
\end{equation}
\textcolor{black}{which can be expanded into its vector inner product form: }
\begin{equation}
f_{\text {terrain },  i}=\mathbf{g}\left(Q_i\right)^T\, \mathbf{g}\left(Q_i\right)
\label{terrain_penalty_i_expanded}
\end{equation}

\textcolor{black}{According to the matrix derivative identity $\frac{\partial}{\partial Q_i}\bigl(u^T u\bigr) = 2\, \frac{\partial u}{\partial Q_i}^T\, u$,  and letting $u = \mathbf{g}(Q_i)$,  we have: }
\begin{equation}
\frac{\partial f_{\text {terrain },  i}}{\partial Q_i}
= 2\left(\frac{\partial \mathbf{g}(Q_i)}{\partial Q_i}\right)^T
\mathbf{g}(Q_i)
\end{equation}

\textcolor{black}{The next step is to compute the Jacobian matrix $\frac{\partial \mathbf{g}(Q_i)}{\partial Q_i}$. We define an intermediate variable $\mathbf{y}_i = W^T Q_i + b$,  allowing $\mathbf{g}(Q_i)$ to be written as: }
\begin{equation}
\mathbf{g}(Q_i)
= W\;\bigl(\beta \odot \cos(\underbrace{W^T Q_i + b}_{\mathbf{y}_i})\bigr)
\end{equation}
\textcolor{black}{Applying the chain rule to differentiate yields: }
\begin{equation}
\frac{\partial \mathbf{g}\left(Q_i\right)}{\partial Q_i} =W \frac{\partial}{\partial Q_i}\left[\beta \odot \cos \left(\mathbf{y}_i\right)\right] W^T
\label{gradient_g_1}
\end{equation}

\begin{equation}
\begin{split}
=\, W\;{%
  \setlength{\arraycolsep}{1pt}
  \begin{pmatrix}
    \beta_1\bigl(-\sin y_{i1}\bigr) & 0 & \cdots & 0\\
    0 & \beta_2\bigl(-\sin y_{i2}\bigr) & \cdots & 0\\
    \vdots & \vdots & \ddots & \vdots\\
    0 & 0 & \cdots & \beta_D\bigl(-\sin y_{iD}\bigr)
  \end{pmatrix}%
}\;W^T
\end{split}
\label{gradient_g_2_explicit}
\end{equation}

\begin{equation}
 = W\left(\beta \odot\left[-\sin \left(\mathbf{y}_i\right)\right]\right) W^T
\label{gradient_g_2}
\end{equation}

\textcolor{black}{Finally,  we combine all the terms: }
\begin{equation}
\frac{\partial f_{\text {terrain },  i}}{\partial Q_i} =2\left[W\left(\beta \odot\left[-\sin \left(\mathbf{y}_i\right)\right]\right) W^T\right]^T \mathbf{g}\left(Q_i\right)
\label{gradient_terrain_1}
\end{equation}
\textcolor{black}{Since the matrix $W\text{diag}(\cdot)W^T$ is symmetric,  its transpose is equal to itself. Thus: }
\begin{equation}
 =2 W\left(\beta \odot\left[-\sin \left(\mathbf{y}_i\right)\right]\right) W^T \mathbf{g}\left(Q_i\right)
\label{gradient_terrain_2}
\end{equation}
\textcolor{black}{Substituting the expression for $\mathbf{g}(Q_i)$ and simplifying using matrix properties,  we arrive at the final closed-form solution: }


\begin{equation}\label{gradient_terrain_3}
\begin{aligned}
  &= -2\, W
    {\setlength{\arraycolsep}{1pt}
      \begin{pmatrix}
        \beta_1\sin(y_{i1}) & 0                      & \cdots & 0 \\
        0                      & \beta_2\sin(y_{i2}) & \cdots & 0 \\
        \vdots                 & \vdots                 & \ddots & \vdots \\
        0                      & 0                      & \cdots & \beta_D\sin(y_{iD})
      \end{pmatrix}
    } W^T \\[6pt]
  &\quad\times W
    {\setlength{\arraycolsep}{1pt}
      \begin{pmatrix}
        \beta_1\cos(y_{i1}) & 0                      & \cdots & 0 \\
        0                      & \beta_2\cos(y_{i2}) & \cdots & 0 \\
        \vdots                 & \vdots                 & \ddots & \vdots \\
        0                      & 0                      & \cdots & \beta_D\cos(y_{iD})
      \end{pmatrix}
    } \\[4pt]
  &= -2\, W\bigl(\beta \odot \sin(\mathbf{y}_i)\bigr)\, W^T\, W\bigl(\beta \odot \cos(\mathbf{y}_i)\bigr)
\end{aligned}
\end{equation}

\textcolor{black}{Thus,  we have successfully derived the analytical gradient of the terrain penalty term with respect to the trajectory control points. This closed-form matrix expression can be computed efficiently and provided directly to a nonlinear optimizer,  thereby forming the mathematical core of our terrain-aware trajectory planning framework. This result fully demonstrates the immense potential of the RMRP method in bridging perception and planning to achieve high-performance autonomous navigation in complex scenarios.}



\begin{figure}[h]
	\centering 
	\includegraphics[width=0.5\textwidth]{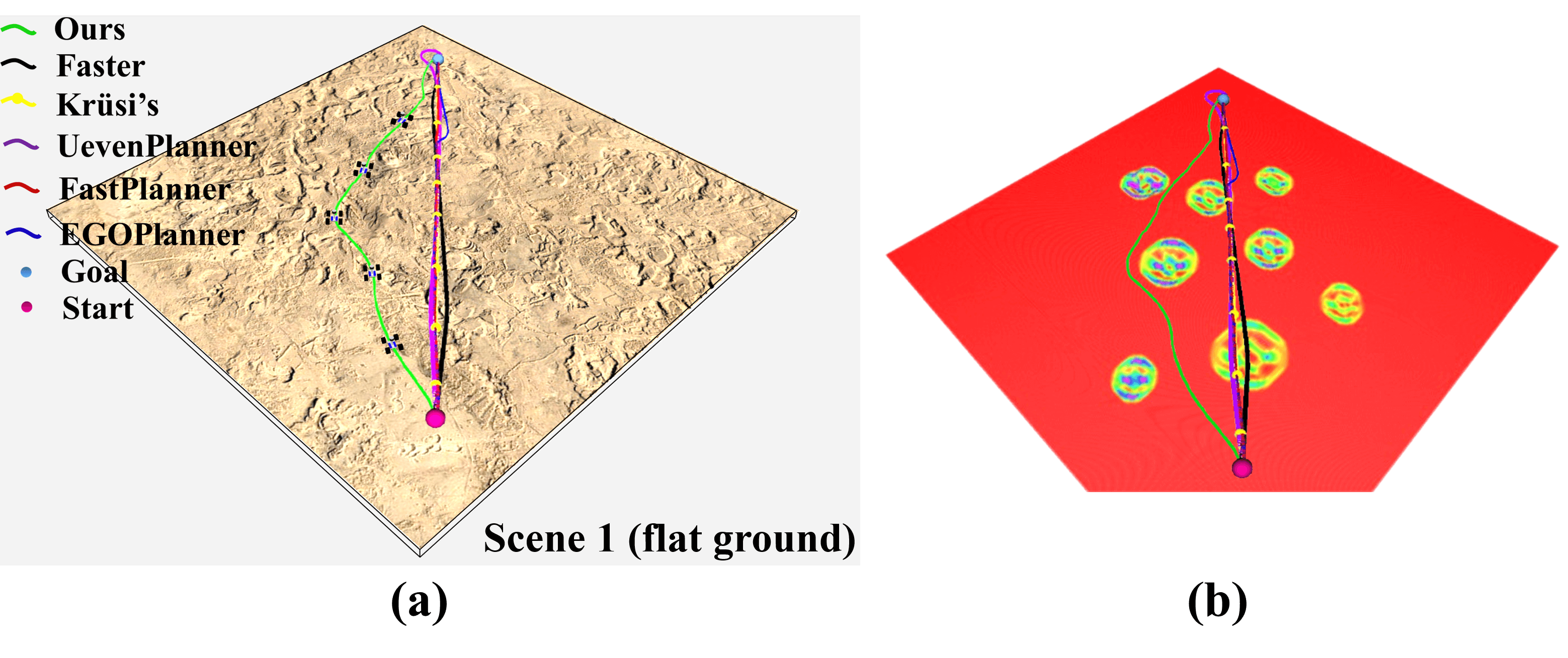   }
	\caption{\textcolor{black}{Experiment I (Zero Obstacle-Avoidance Weights):  With disabled obstacle avoidance,  conventional planners (except Ours and UnevenPlanner) ignored obstacles/depressions,  taking the shortest path. UnevenPlanner inadequately avoided shallow pits. Conversely,  our approach accurately detected and avoided all depressions while maintaining smoothness. (a) Gazebo simulation environment; (b) Corresponding elevation color display. See also \textcolor{black}{Fig. 19}.}} 
	\label{car_contrast1and2}%
\end{figure}

\begin{figure}[h]
	\centering 
	\includegraphics[width=0.5\textwidth]{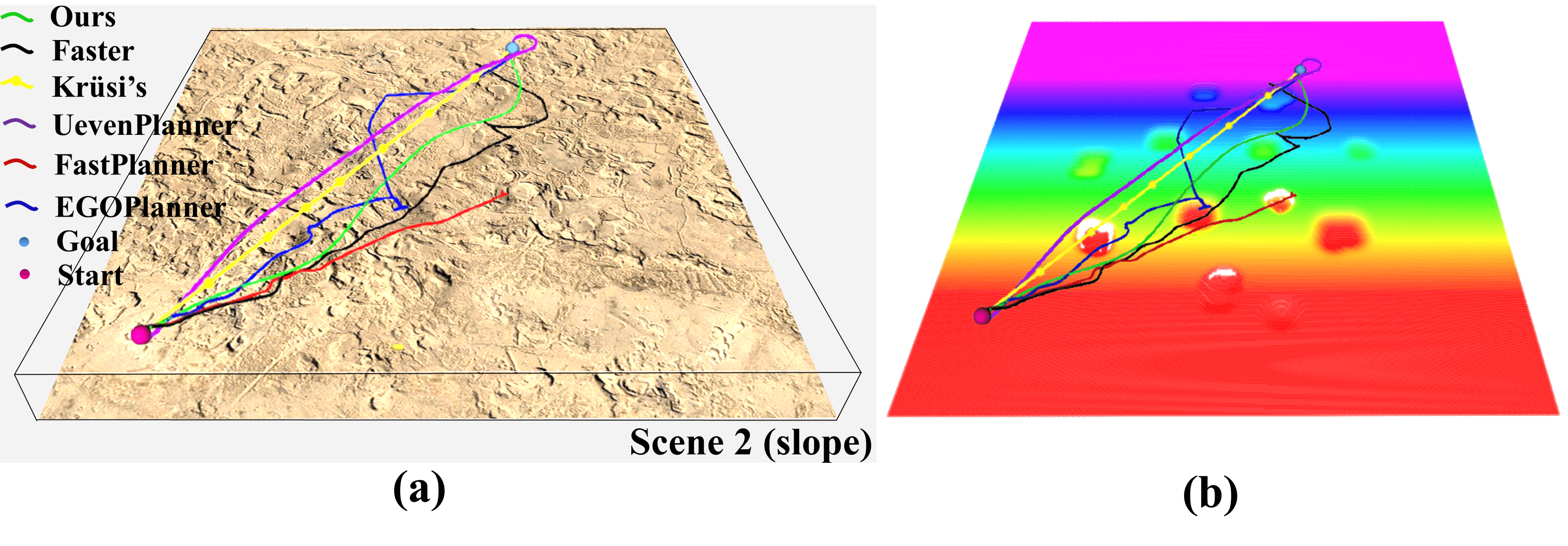   }
	\caption{\textcolor{black}{Experiment II (Default Weights):  Conventional planners,  reverting to default settings,  treated the ground as an obstacle,  hindering progress and failing to avoid depressions. UnevenPlanner showed limited depression avoidance,  still traversing moderately deep pits. Conversely,  our method consistently avoided depressions,  generating smooth,  efficient trajectories. In both scenarios,  our approach reliably avoided depressions while maintaining high trajectory quality,  outperforming conventional planners and UnevenPlanner.}} 
	\label{car_contrast3and4}%
\end{figure}

\vspace{3pt} 

\begingroup
  \setlength{\tabcolsep}{1pt}
\begin{table}[htbp]
  \centering
  \scriptsize                
  \begin{threeparttable}
  \caption{Comparison of planning methods at different map resolutions}
  \label{tab: planning_comparison}
  \begin{tabular}{llrrr}
    \toprule
    Resolution & Method            & Total time (ms) & Path length (m) & Nodes visited \\
    \midrule
    0.20\, m    & Kinodynamic A*    & 31.76           & 52.87           & 27, 704        \\
               & Ours              & \textbf{7.12}            & \textbf{47.34}           & 19, 230        \\
               & informed RRT*     & \textbf{MAX}2000.00         & 63.19           & 20, 537        \\
               & GUILD RRT*        & 1450.00         & 59.62           & \textbf{MAX}30, 000        \\
               & informed BRRT*    & \textbf{MAX}2000.00         & 59.17           & \textbf{9, 763}         \\
    \addlinespace
    0.15\, m    & Kinodynamic A*    & 45.95  & 67.92           & 36, 641 \\
               & Ours           & \textbf{6.72}  & \textbf{49.16}  & 19, 305 \\
               & informed RRT*     & \textbf{MAX}2000.00& 61.24           & 18, 984 \\
               & GUILD RRT*        & 1500.00& 68.72           & \textbf{MAX}30, 000 \\
               & informed BRRT*    & \textbf{MAX}2000.00& 64.21  & \textbf{12, 398} \\
    \addlinespace
    0.10\, m    & Kinodynamic A*    & 51.32  & 60.09           & 39, 872 \\
               & Ours           & \textbf{5.01}  & \textbf{48.77}  & 17, 987 \\
               & informed RRT*     & \textbf{MAX}2000.00& 64.16           & 18, 606 \\
               & GUILD RRT*        & 1600.00& 63.70           & \textbf{MAX}30, 000 \\
               & informed BRRT*    & \textbf{MAX}2000.00& 63.13           & \textbf{9, 911}  \\
    \bottomrule
    \end{tabular}
    \begin{tablenotes}\scriptsize
      \item \textbf{MAX}2000.00 denotes the maximum search time of 2000 ms.
      \item \textbf{MAX}30, 000 denotes the maximum number of nodes (30, 000) visited.
    \end{tablenotes}
  \end{threeparttable}
\end{table}
\endgroup

\section{Experiment and Results}

\subsection{Experimental Setup for UAV Flight Tests}

The Perception-aware trajectory planning framework is summarized in \textcolor{black}{Fig.~\ref{Framework}}. We verify the performance of our method RPAPTR in different cluttered environments. We set the regularization coefficient \( \alpha \) to 0.01 to reduce model complexity and avoid overfitting. The model weight parameters \(\beta\) are quickly solved by AdamW. In order to improve the efficiency of online solving while ensuring accuracy,  the mapping dimension \( M \) of the parameter map model is set to 100.

\subsection{UAV Front-end:  Kinodynamic Search}
We are conducting path search efficiency evaluation in a large-scale scene filled with complex obstacles. Undoubtedly,  resolution plays a crucial role in the efficiency and quality of path searching,  so we compared and evaluated the methods under three different resolutions of 0.2,  0.15,  and 0.1. We compare \textcolor{black}{our method} with classic and state-of-the-art methods:  Informed RRT* \cite{InformedRRTStar},  GUILD sampling RRT* \cite{GUILD},  IB-RRT* \cite{IBRRTStar},  and kinodynamic A* \cite{KinodynamicAStar}. \textcolor{black}{The qualitative difference in the resulting path smoothness and efficiency across the different resolutions is immediately apparent in \textcolor{black}{Fig.~\ref{front_end_multi_resolution_contrast}}.}

Although sampling-based methods (RRT*) do not require map construction and can find feasible solutions,  they have strong randomness and high planning time. Unlike sampling-based methods,  classic path searching methods (kinodynamic A*). They need to build a occpancy map and perform heuristic searches based on Euclidean distance on this occupancy grid map. There are two main sources that affect search efficiency:  one is the time taken to construct the occupancy map,  and the other is the selection of heuristic functions. First,  the popular uniform grid map \cite{zhou2019robust} is compared with our linear parametric map. Mapping time,  access time,  and memory consumption of the grid map are crucial for path searching in large-scale environments. Thus,  mapping time,  access time,  and memory consumption are presented in \textcolor{black}{Table I}.  The mapping time of RMRP is more efficient than the Bayesian filtering. And Bayesian filtering needs to keep the probability values to determine the occupation status,  which brings challenges due to high computation and storage consumption in large-scale environments. Unlike Bayesian methods,  our linear parameter map only needs to update and store some limited model parameters,  making its storage more efficient,  especially for large-scale environments. The novel parametric maps can be easily accessed by substituting $ \bm{g}(\bm{W}x+ b) $ into \eqref{grid_MAP_MDOEL},  giving rise to a \textcolor{black}{shorter} access time.

\begin{figure}[t]
	\centering 
	\includegraphics[width=0.48\textwidth]{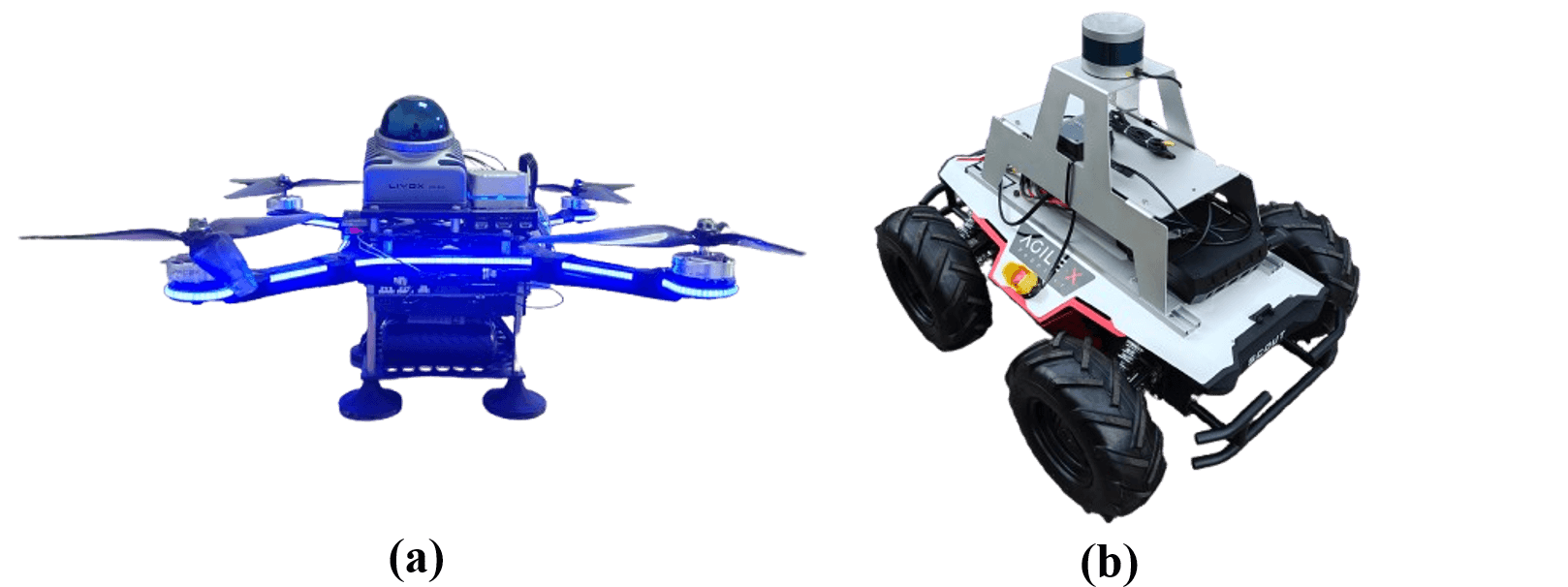}
	\caption{ (a) Experimental UAV Platform. (b) Experimental UGV Platform. 
}
	\label{real_tool}
\end{figure}

Based on the above occupied grid map model,  the occupancy gradient can be derived and used as a novel cost function,  which can refine initial trajectories from planners like Kinodynamic A* for significantly improved safety and smoothness. Thus,  as seen in \textcolor{black}{Table I},  the results showed that our method performs better in search time,  path length,  and other metrics.

\begin{table}[htbp]
  \centering
  \setlength{\tabcolsep}{2pt} 
    \scriptsize                
  \caption{Performance comparison at resolution 0.20\, m}
  \label{tab: planning_res020}
  \begin{tabular}{lrrrrr}
    \toprule
    Planner method   & $t$ (s) & Length (m) & $t_{\mathrm{plan}}$ (s) & Success (\%) & Energy ($\mathrm{m^2/s^3}$) \\
    \midrule
    Fast‐Planner     & 25.18   & 56.15      & 2.51                    & 78           & 906.4               \\
    Faster           & 27.40   & 59.84      & 3.10                    & 71           & 949.2               \\
    EGO‐Planner      & 26.32   & 61.27      & 1.52                    & 82           & 934.8               \\
    \textbf{Ours}    & \textbf{22.57} & \textbf{47.37} & \textbf{1.08}      & \textbf{87}  & \textbf{861.7}               \\
    \bottomrule
  \end{tabular}
\end{table}

\vspace{-12pt}

\begin{table}[htbp]
  \centering
  \setlength{\tabcolsep}{2pt} 
    \scriptsize                
  \caption{Performance comparison at resolution 0.15\, m}
  \label{tab: planning_res015}
  \begin{tabular}{lrrrrr}
    \toprule
    Planner method   & $t$ (s) & Length (m) & $t_{\mathrm{plan}}$ (s) & Success (\%) & Energy ($\mathrm{m^2/s^3}$) \\
    \midrule
    Fast‐Planner     & 27.05   & 60.14      & 2.68                    & 81           & 895.3               \\
    Faster           & 29.86   & 63.18      & 3.26                    & 74           & 942.6               \\
    EGO‐Planner      & 28.64   & 63.13      & 1.75                    & 84           & 914.8               \\
    \textbf{Ours}    & \textbf{23.35} & \textbf{49.28} & \textbf{1.15}      & \textbf{90}  & \textbf{874.1}               \\
    \bottomrule
  \end{tabular}
\end{table}

\vspace{-12pt}

\begin{table}[htbp]
  \centering
  \setlength{\tabcolsep}{2pt} 
    \scriptsize                
  \caption{Performance comparison at resolution 0.10\, m}
  \label{tab: planning_res010}
  \begin{tabular}{lrrrrr}
    \toprule
    Planner method   & $t$ (s) & Length (m) & $t_{\mathrm{plan}}$ (s) & Success (\%) & Energy ($\mathrm{m^2/s^3}$) \\
    \midrule
    Fast‐Planner     & 26.92   & 58.76      & 2.71                    & 84           & 882.5               \\
    Faster           & 30.54   & 67.46      & 3.35                    & 77           & 961.0               \\
    EGO‐Planner      & 29.41   & 61.13      & 1.85                    & 87           & 905.4               \\
    \textbf{Ours}    & \textbf{22.19} & \textbf{48.62} & \textbf{1.02}      & \textbf{92}  & \textbf{858.9}               \\
    \bottomrule
  \end{tabular}
\end{table}

\vspace{-15pt}

\subsection{UAV Back-end:  trajectory optimization}
We compare the proposed RPATR method with three state-of-the-art approaches:  Fast-Planner \cite{zhou2019robust},  EGO-Planner \cite{zhou2020ego},  and Faster \cite{tordesillas2021faster}. All methods utilize their default recommended parameter settings,  including identical controller parameters and consistent maximum velocity/acceleration constraints for trajectory tracking. While traditional methods require computationally intensive ESDF map calculations at appropriate resolutions for trajectory optimization (which relies on distance and gradient values),  our approach employs a closed-form ESDF solution,  enabling efficient computation without interpolation. This efficiency facilitates real-time online trajectory optimization in the back-end.As shown in \textcolor{black}{Fig. \textcolor{black}{15}},  we evaluated the trajectory optimization using three different resolutions. Comparative results for ESDF update time,  trajectory length,  and planning time are presented in \textcolor{black}{Table II-IV} and \textcolor{black}{Fig. 8}. The data demonstrates that RPATR achieves superior performance in both ESDF update and planning efficiency.

\begin{figure}[t]
	\centering 
	\includegraphics[width=0.38\textwidth]{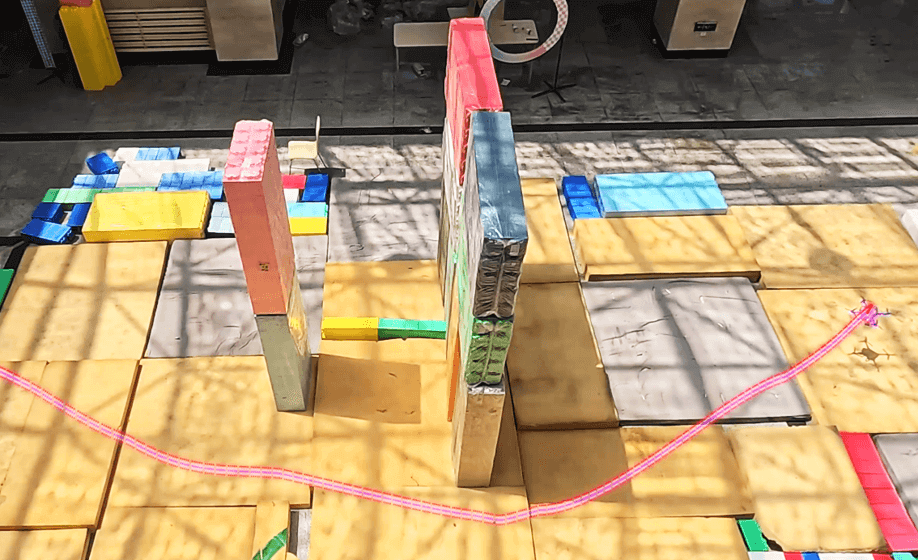}
	\caption{\textcolor{black}{Trajectory of UAV in Scenario 1 (Indoor Experiment).}
}
	\label{UAV_indoor1}
\end{figure}

\begin{figure}[t]
	\centering 
	\includegraphics[width=0.38\textwidth]{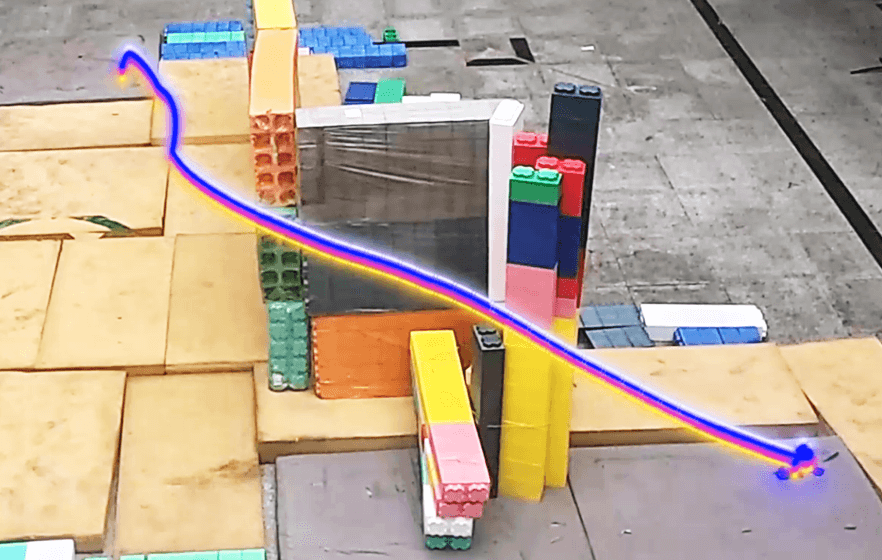}
	\caption{\textcolor{black}{Trajectory of UAV in Scenario 2 (Indoor Experiment).}
}
	\label{UAV_indoor2}
\end{figure}

\vspace{-10pt}

\subsection{Real-World UAV Experiments}

\subsubsection{\textcolor{black}{Implementation Details}}

\textcolor{black}{To validate the end-to-end performance of the proposed RMRP framework in a real-world physical environment,  we conducted autonomous flight experiments on an NVIDIA Jetson Xavier NX platform.\textsuperscript{1} The experimental UAV platform used is shown in \textcolor{black}{Fig. 20(a)}. Equipped with a Livox Mid360 LiDAR, \textsuperscript{2} the platform executed the complete perception and planning pipeline in real-time within a 15m×25m×4m indoor environment. The core planner utilizes a Kinodynamic A* algorithm deeply coupled with the RMRP model,  leveraging its predictive capabilities for real-time obstacle completion in sensor blind spots (Figs.~\ref{UAV_indoor1}-\ref{drone_scene2_predict}). The experimental results demonstrate that the framework supports the UAV in performing aggressive dynamic obstacle avoidance and waypoint-following tasks at speeds of up to 6 m/s,  validating its capability for efficient and safe end-to-end autonomous navigation in complex scenarios.}

\ifnum\value{page}=14
    \noindent\begin{flushleft}
      \footnotesize\textsuperscript{1}\url{https: //www.nvidia.com/en-us/autonomous-machines/embedded-systems/jetson-xavier-series/}
    \end{flushleft}
\fi
\vspace{-12pt}
\ifnum\value{page}=14
    \noindent\begin{flushleft}
      \footnotesize\textsuperscript{2}\url{https: //www.livoxtech.com/mid-360}
    \end{flushleft}
\fi



\subsubsection{\textcolor{black}{Perception-Aware Trajectory Planning for UAVs}}
We present two complex indoor scenarios to demonstrate the advantages of our perception-awareness planning method. \textcolor{black}{A conceptual illustration comparing a traditional reactive planner with our proactive,  perception-aware approach is shown in \textcolor{black}{Fig. 9}.}We compare the proposed perception-awareness planning method with the fast planner method. Each planner is tested 5 times in both scenes and we record the number of successful flights. The maximum flight velocity is set as 5m/s,  the maximum flight acceleration is set as 4 $m/s^2$.

In the first scenario,  two obstacles are placed in front of the plane,  as shown in \textcolor{black}{Fig. 21}. A small cube is in front,  and a long and large one is behind. The final target of the plane is 4 meters behind the center point of the second large obstacle. Without perception-aware planning methods lack the ability to perceive the occupancy state of occluded regions (marked in red) in advance. Consequently,  only when the UAV arrivate the first obstacle does it detect the occupancy of these occluded areas. This late detection forces computationally expensive emergency replanning to locally optimize the trajectory away from the obstacle while still attempting to reach the goal. In contrast,  our  perception-aware planning method overcomes this limitation. By leveraging offline model training combined with real-time online perception,  the RPATR enables predictive mapping of occluded regions behind obstacles. As is seen on \textcolor{black}{Fig.11-12,  21-22},  this predictive capability facilitates proactive global trajectory optimization before the UAV reaches critical proximity to obstacles. 

\begin{figure*}[!t]
  \centering 
  \includegraphics[width=\textwidth]{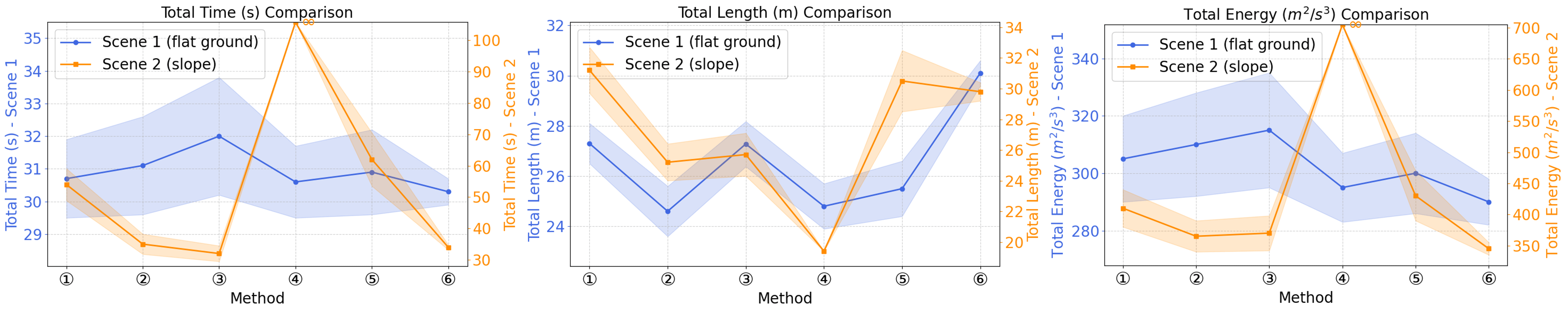}
  \caption{\textcolor{black}{Performance comparison of different planners in two scenarios.
  The plots compare our proposed method against five others based on (a) total time ($t_{\mathrm{tot}}$), 
  (b) total path length ($l_{\mathrm{tot}}$),  and (c) total energy ($E_{\mathrm{eng}}$).
  Solid lines represent the mean performance over multiple trials,  while the shaded areas indicate
  the standard deviation,  reflecting the stability of each method. The $\infty$ symbol indicates that
  the Fast-Planner failed to complete the task in the slope scenario,  with its data point placed at
  the top of the axis for visual representation.}}
  \label{car_compare_three}
\end{figure*}

\begingroup
  \setlength{\tabcolsep}{3pt} 
  \begin{table}[htbp]
    \centering
    \scriptsize                
    \caption{Comparison of planners in Scene 1 (flat) and Scene 2 (slope),  with mean ± standard deviation.}
    \label{tab: scene_comparison_std}
    \begin{tabular}{lrrrrc}
      \toprule
      Method          & $n_{\mathrm{pit}}$ & $t_{\mathrm{tot}}$ (s) & $l_{\mathrm{tot}}$ (m)
                      & $E_{\mathrm{eng}}$ ($\mathrm{m^2/s^3}$) & $online_{\mathrm{map}}$ \\
      \midrule
      \multicolumn{6}{l}{\textit{Scene 1 (flat ground)}} \\
      \ding{172}Faster          & 2  & 30.7 ± 1.2   & 27.30 ± 0.8   & 305 ± 15       & \cmark \\
      \ding{173}Krüsi’s            & 2  & 31.1 ± 1.5   & 24.60 ± 1.0   & 310 ± 18       & \xmark    \\
      \ding{174}Uneven Planner  & 2  & 32.0 ± 1.8   & 27.28 ± 0.9   & 315 ± 20       & \xmark    \\
      \ding{175}Fast-Planner    & 2  & 30.6 ± 1.1   & 24.80 ± 0.9   & 295 ± 12       & \cmark \\
      \ding{176}EGO-Planner     & 2  & 30.9 ± 1.3   & 25.50 ± 1.1   & 300 ± 14       & \cmark \\
      \textbf{\ding{177}Ours}   & \textbf{0}  & \textbf{30.3 ± 0.4} & \textbf{30.10 ± 0.5}
                      & \textbf{290 ± 8}  & \cmark \\
      \addlinespace
      \multicolumn{6}{l}{\textit{Scene 2 (slope)}} \\
      \ding{172}Faster          & 3  & 54.0 ± 5.1   & 31.20 ± 1.5   & 410 ± 30       & \cmark \\
      \ding{173}Krüsi’s            & 2  & 35.0 ± 3.2   & 25.20 ± 1.2   & 365 ± 25       & \xmark    \\
      \ding{174}Uneven Planner  & 2  & 32.0 ± 2.5   & 25.70 ± 1.4   & 370 ± 28       & \xmark    \\
      \ding{175}Fast-Planner    & 1  & $\infty$     & 19.40         & $\infty$       & \cmark \\
      \ding{176}EGO-Planner     & 4  & 62.0 ± 8.5   & 30.50 ± 2.0   & 430 ± 40       & \cmark \\
      \textbf{\ding{177}Ours}   & \textbf{0}  & \textbf{34.0 ± 0.8} & \textbf{29.80 ± 0.6}
                      & \textbf{345 ± 10} & \cmark \\
      \bottomrule
    \end{tabular}
  \end{table}
\endgroup


\subsubsection{\textcolor{black}{Fully autonomous flight}}
\textcolor{black}{To further validate our method, } we conducted fully autonomous flight tests within a large-scale forest environment. \textcolor{black}{This real-world outdoor scene,  approximately 30m×100m in size,  is characterized by densely packed,  randomly distributed obstacles (e.g.,  tree trunks and branches) and large-scale,  unstructured features.} The substantial volume of irregular point cloud data acquired by onboard sensors presents significant challenges for constructing high-quality maps and executing efficient real-time planning under constrained computational resources. The UAV successfully navigates through complex obstacles,  including tree trunks and foliage,  achieving agile flight (\textcolor{black}{Fig. 24}). During these tests,  the UAV reached a maximum velocity of \textcolor{black}{8 m/s} and maintained an average velocity of \textcolor{black}{6 m/s}. The experiment results demonstrate the capability of the proposed method to enable robust autonomous flight in a large-scale and complex field environment.



\textcolor{black}{\subsection{Perception-Aware Trajectory Planning for UGVs}}

\subsubsection{Experimental Setup and Overall Performance}

\textcolor{black}{In our experiments,  we utilize the SCOUT 2.0 platform, \textsuperscript{3} a robust four-wheel differential-drive all-terrain mobile robot developed by AgileX Robotics. This experimental UGV platform is shown in \textcolor{black}{Fig. 20(b)}. The system is further customized using the Robot Operating System (ROS) to support advanced perception and motion planning capabilities. For computational support,  the platform is equipped with an NVIDIA GeForce RTX 2060 GPU,  providing sufficient onboard processing power for real-time SLAM and trajectory computation tasks.
We employ A-LOAM as our SLAM framework,  which offers high-accuracy pose estimation in outdoor environments. The perception module is supported by a Velodyne VLP-16 LiDAR sensor, \textsuperscript{4} providing dense 3D point cloud data at a refresh rate of 10 Hz. This sensor configuration ensures robust mapping performance and reliable localization under varying environmental conditions.}

\ifnum\value{page}=15
    \noindent\begin{flushleft}
      \footnotesize\textsuperscript{3}\url{https: //global.agilex.ai/products/scout-2-0}
    \end{flushleft}
\fi
\vspace{-12pt}
\ifnum\value{page}=15
    \noindent\begin{flushleft}
      \footnotesize\textsuperscript{4}\url{https: //ouster.com/products/hardware/vlp-16}
    \end{flushleft}
\fi

\textcolor{black}{To quantitatively evaluate the effectiveness and robustness of our proposed RMRP-based terrain-aware trajectory planning framework,  we conducted comprehensive comparisons against five state-of-the-art baseline methods (Faster \cite{tordesillas2021faster},  Krüsi’s \cite{krusi2017driving},  Uneven Planner \cite{xu2021autonomous},  Fast-Planner \cite{zhou2019robust},  and EGO-Planner \cite{zhou2020ego}) in two representative simulation scenarios. Scenario 1,  which tests performance on flat ground with a pit,  is visually detailed in \textcolor{black}{Fig. 18}. Scenario 2,  featuring a more complex slope with a pit,  is shown in \textcolor{black}{Fig. 19}. The evaluation metrics cover safety ($n_{\mathrm{pit}}$),  efficiency ($t_{\mathrm{tot}}$,  $E_{\mathrm{eng}}$),  path quality ($l_{\mathrm{tot}}$),  and online adaptability ($online_{\mathrm{map}}$). The detailed quantitative results are presented in \textcolor{black}{Table~\ref{tab: scene_comparison_std}},  and a visual comparison of key performance metrics is provided in \textcolor{black}{Fig.~\ref{car_compare_three}}. The results clearly indicate that our proposed method is the only planner to achieve zero pit collisions ($n_{\mathrm{pit}}=0$) in all tests,  while also demonstrating a comprehensive advantage in runtime efficiency and performance stability. The following sections provide an in-depth analysis of the root causes behind these results.}

\subsubsection{Analysis of Baseline Method Failure Modes}

\textcolor{black}{An analysis of the failures observed in the baseline methods reveals common limitations in existing techniques for complex ground navigation,  which primarily stem from a fundamental mismatch between their core design philosophies and physical reality.}

\textcolor{black}{\textbf{Fundamental Mismatch in Design Philosophy:  The 3D Free-Space Assumption: } For planners designed for UAVs,  such as Faster,  Fast-Planner,  and EGO-Planner,  failure stems from a fundamental design assumption:  the robot is a point mass that can move freely in 3D space,  and its primary task is to avoid collisions with geometric obstacles. When applied to UGVs,  their world models (e.g.,  ESDF or convex safe corridors) erroneously classify the unobstructed area above a pit as safe free space. Concurrently,  their optimization objective of seeking the shortest and smoothest paths actively guides the trajectory through this aerial route across the pit. Consequently,  these methods exhibit high $n_{\mathrm{pit}}$ values in \textcolor{black}{Table~\ref{tab: scene_comparison_std}},  as they lack an understanding of the core concept of terrain traversability and the necessary ground constraint.}


\begin{figure*}[t]
	\centering 
	\includegraphics[width=1.0\textwidth] {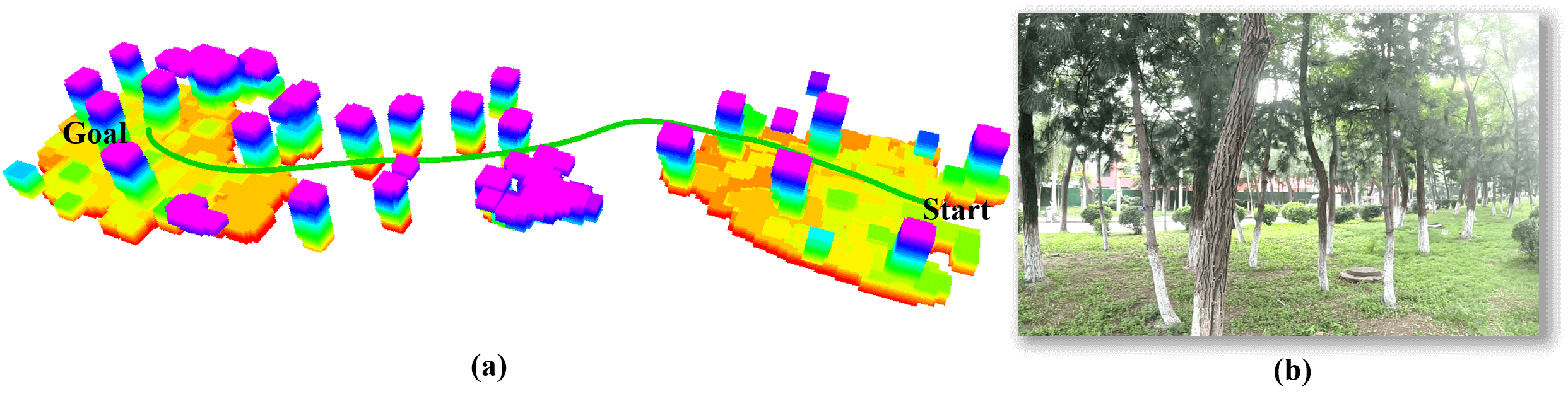}
	\caption{\textcolor{black}{showcases the outdoor flight experiment within a forest environment. (a) illustrates the UAV trajectory and the online-generated forest map,  represented as a closed-form ESDF,  with elevation indicated by color. (b) depicts the real-world scene of the autonomous fast flight within the dense forest.}
}
	\label{outdoor_real_123}
\end{figure*}

\begin{figure*}[t]
	\centering 
	\includegraphics[width=1.0\textwidth] {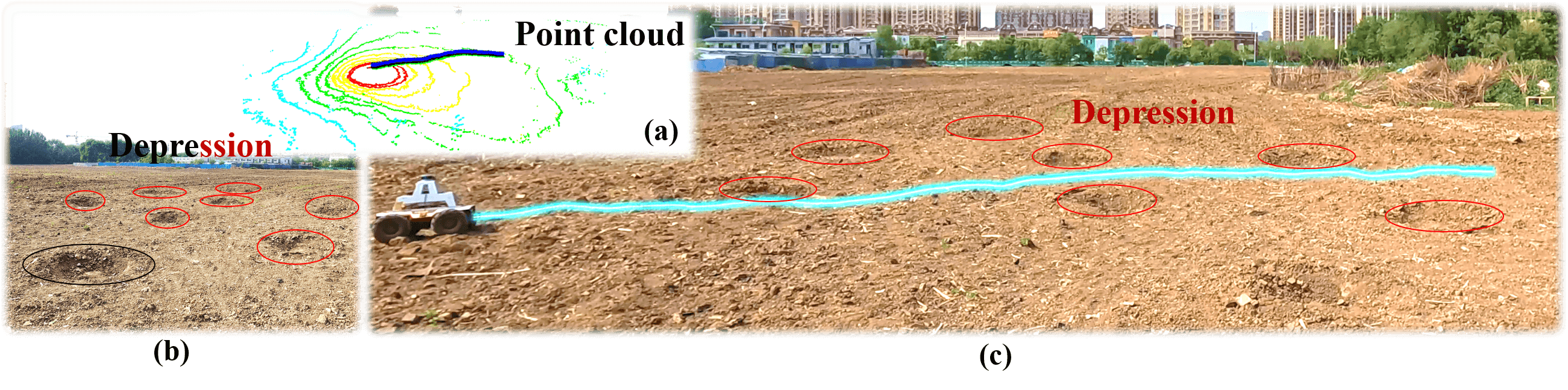}
	\caption{\textcolor{black}{This figure showcases the UGV's real-world experiment in a terrain environment. (a) presents the raw point cloud data acquired by the Mid360 LiDAR; (b) shows the side-view profiles of multiple pits within the terrain,  which are annotated with different colors; (c) depicts the UGV successfully circumventing these pits. The red circles indicate the locations of the pits,  corresponding one-to-one with the profiles in (b),  while the blue curve represents the UGV's safe avoidance trajectory.}
}
	\label{car_real1_another}
\end{figure*}

\textcolor{black}{\textbf{Architectural Limitations of Dedicated Ground Planners: } Even planners designed specifically for ground navigation are limited by inherent architectural flaws when encountering specific hazards. The SE(2)-to-SE(3) mapping and ellipsoid-fitting strategy employed by Uneven Planner leads to an over-simplification of the terrain,  resulting in the loss of critical geometric information about the pit (e.g.,  depth,  steep edges). Its coarse terrain constraint mechanism is also insufficient to identify such specific hazards. While Krüsi’s Method ensures high safety through a dedicated terrain assessment module,  its complex three-stage RRT → RRT* → Smoothing process results in prohibitive computational costs (as reflected by the long ttot in \textcolor{black}{Table~\ref{tab: scene_comparison_std}}) and poor trajectory quality,  demonstrating an extreme trade-off between safety and computational efficiency that makes it impractical for real-time tasks.}

\subsubsection{The Advantage of the RMRP Framework}

\textcolor{black}{In stark contrast to the limitations of the baseline methods,  the success of our approach stems from its innovative framework that deeply integrates perception and planning.
}

\textcolor{black}{\textbf{Continuous and Differentiable Terrain Model: } The cornerstone of our method is the continuous and fully differentiable terrain model learned via RMRP,  as depicted in \textcolor{black}{Fig.~\ref{car_regression}}. This model accurately represents complex terrain geometry,  including pits,  overcoming the discreteness and simplification issues of traditional models. \textbf{Risk Quantification via Analytical Gradients: }  Leveraging the differentiability of this model,  we derive the analytical gradient of the terrain elevation. This gradient is used to formulate a targeted terrain cost function (Eq. \ref{terrain_penalty2}),  which directly and accurately translates the geometric features of high-risk regions,  such as pit edges,  into a continuous mathematical cost understood by the optimizer. \textbf{Experimental Validation: }  The effectiveness of this mechanism is fully validated by our experiments. As shown in \textcolor{black}{Fig. 17},  it is precisely this gradient-based cost term that guides the optimizer to actively plan a smooth and safe trajectory around the hazardous pit. This directly explains why our method achieves a perfect safety record of $n_{\mathrm{pit}}=0$ in \textcolor{black}{Table~\ref{tab: scene_comparison_std}},  and also obtains the lowest energy consumption and highest stability (as indicated by the minimal standard deviation in \textcolor{black}{Fig.~\ref{car_compare_three}}) due to more reasonable path planning.}

\textcolor{black}{Finally,  to validate that the strong performance observed in simulation translates to real-world scenarios,  we deployed the UGV platform in a physical terrain environment with multiple natural depressions. The results of this real-world validation are presented in \textcolor{black}{Fig. 25}. As shown,  the system successfully utilized its onboard sensors to build a terrain model,  identify the hazardous pits,  and generate a smooth,  safe trajectory to circumvent them,  confirming the practical applicability and robustness of our proposed framework.}


\subsection{Efficient Implementation of Learning Modules}
\textcolor{black}{
Our learning framework is engineered for real-time performance through deep integration with NVIDIA CUDA.\textsuperscript{5} We accelerate core computations using custom-written kernels,  leveraging mixed-precision (FP16)\textsuperscript{6} training to harness modern GPU tensor cores for significant reductions in latency and memory footprint. This is complemented by a dedicated,  GPU-native implementation of the AdamW optimizer to minimize data transfer overhead during training. The framework supports both full,  ab-initio training for initial model generation and efficient incremental learning,  enabling rapid in-situ adaptation as new data is perceived.
}

\ifnum\value{page}=17
    \noindent\begin{flushleft}
      \footnotesize\textsuperscript{5}\url{https: //developer.nvidia.com/cuda-toolkit}
    \end{flushleft}
\fi
\vspace{-12pt}
\ifnum\value{page}=17
    \noindent\begin{flushleft}
      \footnotesize\textsuperscript{6}\url{https: //standards.ieee.org/standard/754-2019.html}
    \end{flushleft}
\fi






\section{\textcolor{black}{Conclusion}}

We propose a novel lightweight navigation mapping approach via \textcolor{black}{RMRP}. For \textcolor{black}{UAVs},  grid maps and \textcolor{black}{ESDF} maps are constructed as a unified linear parametric map. Leveraging the continuous linear parametric model of the grid map,  occupancy status can be queried rapidly and the gradient of the occupancy degree can be efficiently obtained to accelerate kinodynamic path searching efficiency in the front-end. Furthermore,  the learned linear parametric model yields a closed-form ESDF map,  enabling high-accuracy and effective trajectory optimization in the back-end. Crucially,  the linear parametric map is globally continuous and facilitates online prediction of occupancy in unobserved regions,  supporting perception-aware trajectory planning in complex environments. For \textcolor{black}{UGVs},  we employ the linear parametric model to characterize terrain maps. Based on this model,  terrain gradients are derived through a closed-form solution,  enabling online perception-aware trajectory planning to circumvent large holes. We validate our linear parametric maps and perception-aware trajectory planning methods across diverse indoor and outdoor scenarios. The results demonstrate that our linear parametric maps achieve superior comprehensive performance in terms of mapping time,  memory consumption,  access time,  and accuracy. Simultaneously,  our perception-aware planning method exhibits enhanced performance across various challenging environments:  it is computationally lightweight and efficient in large-scale scenes,  ensures safe high-speed flight in complex obstacle-filled settings,  and enables safe online avoidance of pits and uneven terrain.

\section*{\textcolor{black}{Appendix}}

\subsection{\textbf{Derivation of the Residual Energy Preservation Theorem}}

\begin{proof}
Let 
\[
  \mathbf{e}_{\mathrm{proj}}
  : = \mathbf{R}\, \mathbf{x}_{\perp}
     - P_{\mathbf{R}\mathcal{S}}\, \mathbf{R}\, \mathbf{x}_{\perp}.
\]
We condition on the high‐probability event
\(\mathcal{E}_{\mathrm{len}}\land\mathcal{E}_{\mathrm{ang}}\land\mathcal{E}_{\mathrm{hw}}\)
(which holds with probability at least \(1-\delta\)).

\paragraph{ \textcolor{black}{Length Preservation}}  
By the \(\varepsilon\)-subspace embedding property (\(\mathcal{E}_{\mathrm{len}}\)), 
\begin{equation}
(1-\varepsilon)\|\mathbf{x}_{\perp}\|
\;\le\;
\|\mathbf{R}\, \mathbf{x}_{\perp}\|
\;\le\;
(1+\varepsilon)\|\mathbf{x}_{\perp}\|
\quad \text{\cite{WoodruffLecture2020, JLNote2009}.} 
\end{equation}

\paragraph{ \textcolor{black}{Angle Leakage}}  
From the Davis–Kahan/Wedin \(\sin\Theta\) theorem (\(\mathcal{E}_{\mathrm{ang}}\)), 
\begin{equation}
  \|P_{\mathcal{S}} - P_{\mathbf{R}\mathcal{S}}\|_{2}
  = \|\sin\Theta(\mathcal{S}, \mathbf{R}\mathcal{S})\|_{2}
  \;\le\;\varepsilon
  \quad \text{\cite{Yu2015, HsuDK16}}. 
\end{equation}
Hence for any \(\mathbf{x}_{\perp}\perp\mathcal{S}\), 
\begin{equation}
\begin{split}
  \|P_{\mathbf{R}\mathcal{S}}\, \mathbf{R}\, \mathbf{x}_{\perp}\|
  &\le \sqrt{2}\, \varepsilon\, \|\mathbf{R}\, \mathbf{x}_{\perp}\| \\[4pt]
  &\le \sqrt{2}\, (1+\varepsilon)\, \varepsilon\, \|\mathbf{x}_{\perp}\| \\[4pt]
  &= C_{1}\, \varepsilon\, \|\mathbf{x}_{\perp}\|,
\end{split}
\quad\text{\cite{MathSE_DKsqrt2}}
\end{equation}

\paragraph{ \textcolor{black}{Hanson–Wright Concentration}}  
By the sparse Hanson–Wright inequality (\(\mathcal{E}_{\mathrm{hw}}\)), 
\begin{equation}
  \Pr\Bigl[\bigl|\|\mathbf{R}\, \mathbf{x}_{\perp}\|^{2}
  - \|\mathbf{x}_{\perp}\|^{2}\bigr|
  > \varepsilon\, \|\mathbf{x}_{\perp}\|^{2}\Bigr]
  \;\le\;\frac{\delta}{2}
  \quad\text{\cite{RudelsonVershynin2013, SparseHW2024, Andoni2024}}.
\end{equation}

\paragraph{ \textcolor{black}{\(\ell_{2}\)‐Norm Bounds}}  
Applying the triangle inequality, 
\begin{equation}
\begin{split}
  \|\mathbf{e}_{\mathrm{proj}}\|
  &\ge \|\mathbf{R}\, \mathbf{x}_{\perp}\|
    - \|P_{\mathbf{R}\mathcal{S}}\, \mathbf{R}\, \mathbf{x}_{\perp}\|,  \\[4pt]
  \|\mathbf{e}_{\mathrm{proj}}\|
  &\le \|\mathbf{R}\, \mathbf{x}_{\perp}\|
    + \|P_{\mathbf{R}\mathcal{S}}\, \mathbf{R}\, \mathbf{x}_{\perp}\|.
\end{split}
\end{equation}

Substituting the results from (1) and (2) and setting \(C_{2}: =1+C_{1}\), 
we obtain
\begin{equation}
  (1 - C_{2}\varepsilon)\|\mathbf{x}_{\perp}\|
  \;\le\;
  \|\mathbf{e}_{\mathrm{proj}}\|
  \;\le\;
  (1 + C_{2}\varepsilon)\|\mathbf{x}_{\perp}\|.
\end{equation}

\paragraph{ \textcolor{black}{Squared‐Norm Bounds}}  
Squaring both sides yields
\begin{equation}
  (1 - C_{2}\varepsilon)^{2}\|\mathbf{x}_{\perp}\|^{2}
  \;\le\;
  \|\mathbf{e}_{\mathrm{proj}}\|^{2}
  \;\le\;
  (1 + C_{2}\varepsilon)^{2}\|\mathbf{x}_{\perp}\|^{2}.
\end{equation}
For \(\varepsilon\le0.3\),  the \(O(\varepsilon^{2})\) terms are negligible, 
so
\begin{equation}
  \bigl|\|\mathbf{e}_{\mathrm{proj}}\|^{2}
        - \|\mathbf{x}_{\perp}\|^{2}\bigr|
  \;\le\;
  2\, C_{2}\, \varepsilon\, \|\mathbf{x}_{\perp}\|^{2}.
\end{equation}

Finally,  combining these bounds under
\(\Pr[\mathcal{E}_{\mathrm{len}}\land\mathcal{E}_{\mathrm{ang}}\land\mathcal{E}_{\mathrm{hw}}]\ge1-\delta\)
via the union bound \cite{Feller1971} completes the proof.
\end{proof}

\bibliographystyle{IEEEtran}
\bibliography{IEEEabrv, TRO}\ 

\vspace{0pt plus 0pt}

\nointerlineskip\vspace{0pt}


\vfill

\end{document}